\documentclass[sigconf]{aamas}

\usepackage{balance} 
\usepackage{algpseudocode}  
\usepackage[ruled, linesnumbered, boxed]{algorithm2e}
\usepackage{parskip}
\usepackage{changepage}

\usepackage{color}
\usepackage{url}            
\usepackage{wrapfig}
\usepackage{caption}
\usepackage{cancel}
\usepackage{graphicx}
\usepackage{subcaption}
\usepackage{epsfig} 
\usepackage{booktabs}
\usepackage{amsmath}
\usepackage{amsthm}
\usepackage[table, dvipsnames, svgnames, x11names]{xcolor} 
\definecolor{ggdarkred}{HTML}{DC4437}
\definecolor{ggblue}{HTML}{4385F5}
\definecolor{gggray}{HTML}{B2B2B2}
\definecolor{gglightblue}{HTML}{A0C2FF}
\definecolor{ggorange}{HTML}{F5B400}
\definecolor{gggreen}{HTML}{109D59}

\usepackage{marvosym}
\definecolor{mygray}{gray}{.9}
\usepackage{pifont}
\usepackage{nicefrac}       
\usepackage{microtype}      
\usepackage[table, dvipsnames, svgnames, x11names]{xcolor}        
\usepackage{colortbl}

\usepackage{multirow}
\usepackage{cite}

\usepackage{enumitem}
\usepackage{graphics} 
\usepackage{svg}
\usepackage{bm}
\usepackage{float} 
\usepackage{stfloats}
\usepackage{adjustbox}
\usepackage{tablefootnote}
\usepackage[colorlinks,
linkcolor=blue,
anchorcolor=blue,
citecolor=blue,
urlcolor=Aqua,
backref=page]{hyperref}

\doi{BENP4894}

\makeatletter
\gdef\@copyrightpermission{
  \vspace{-5pt}
  \begin{minipage}{0.2\columnwidth}
   \href{https://creativecommons.org/licenses/by/4.0/}{\includegraphics[width=0.90\textwidth]{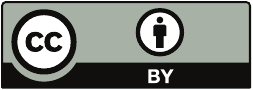}}
  \end{minipage}\hfill
  \begin{minipage}{0.8\columnwidth}
   \href{https://creativecommons.org/licenses/by/4.0/}{This work is licensed under a Creative Commons Attribution International 4.0 License.}
  \end{minipage}
  \vspace{5pt}
}
\makeatother

\setcopyright{ifaamas}
\acmConference[AAMAS '26]{Proc.\@ of the 25th International Conference
on Autonomous Agents and Multiagent Systems (AAMAS 2026)}{May 25 -- 29, 2026}
{Paphos, Cyprus}{C.~Amato, L.~Dennis, V.~Mascardi, J.~Thangarajah (eds.)}
\copyrightyear{2026}
\acmYear{2026}
\acmDOI{}
\acmPrice{}
\acmISBN{}
\acmSubmissionID{422}

\title[Cross-Domain Trajectory Editing (xTED)]{xTED: Cross-Domain Adaptation\\via Diffusion-Based Trajectory Editing}

\author{Haoyi Niu\textsuperscript{*}}
\affiliation{
  \institution{Tsinghua University}
  \city{Beijing}
  \country{China}
  }
  \thanks{\textsuperscript{*}Equal Contribution: \texttt{t6.da.thu@gmail.com, cqm24@mails.tsinghua.edu.cn}.}

\author{Qimao Chen\textsuperscript{*}}
\affiliation{
  \institution{Tsinghua University}
  \city{Beijing}
  \country{China}
  }

\author{Tenglong Liu}
\affiliation{
  \institution{National University of Defense Technology}
  \city{Changsha}
  \country{China}
  }

\author{Jianxiong Li}
\affiliation{
  \institution{Tsinghua University}
  \city{Beijing}
  \country{China}
  }
\vspace{-25pt}
\author{Guyue Zhou}
\affiliation{
  \institution{Tsinghua University}
  \city{Beijing}
  \country{China}
  }

\author{Yi Zhang\textsuperscript{\Letter}}
\affiliation{
  \institution{Tsinghua University}
  \city{Beijing}
  \country{China}
  }
  \thanks{\textsuperscript{\Letter} Corresponding authors: \texttt{\{zhyi, hujm, zhanxianyuan\}@tsinghua.edu.cn}.}

\author{Jianming Hu\textsuperscript{\Letter}}
\affiliation{
  \institution{Tsinghua University}
  \city{Beijing}
  \country{China}
  }

\author{Xianyuan Zhan\textsuperscript{\Letter}}
\affiliation{
  \institution{Tsinghua University}
  \city{Beijing}
  \country{China}
  }

\begin{abstract}
Reusing pre-collected data from different domains is an appealing solution for decision-making tasks, especially when data in the target domain is limited.
Existing cross-domain policy transfer methods mostly aim at learning domain correspondences or corrections to facilitate policy learning, such as learning task/domain-specific discriminators, representations, or policies. This design philosophy often results in heavy model architectures or task/domain-specific modeling, lacking flexibility.
This reality makes us wonder: can we directly bridge the domain gaps universally at the data level, instead of relying on complex downstream cross-domain policy transfer procedures?
In this study, we propose the \textbf{Cross}-Domain \textbf{T}rajectory \textbf{ED}iting (\textbf{xTED}) framework that employs a specially designed diffusion model for cross-domain trajectory adaptation.
Our proposed model architecture effectively captures the intricate dependencies among states, actions, and rewards, as well as the dynamics patterns within target data. 
Edited by adding noises and denoising with the pre-trained diffusion model, source domain trajectories can be transformed to align with target domain properties while preserving original task semantic information. 
This process effectively corrects underlying domain gaps, enhancing state realism and dynamics reliability in source data, and allowing flexible integration with various single-domain and cross-domain downstream policy learning methods.
Despite its simplicity, xTED demonstrates superior performance in extensive simulation and 
\href{https://xted24.github.io/xTED/}{real-robot experiments}. Code is available at this \href{https://github.com/t6-thu/xTED}{repository}.
\end{abstract}

\keywords{Cross-Embodiment, Cross-Domain, Diffusion-Based Editing, Robotics}

\newcommand{\BibTeX}{\rm B\kern-.05em{\sc i\kern-.025em b}\kern-.08em\TeX}

\begin{document}




\maketitle 


	\section{Introduction}
        Solving real-world tasks with reinforcement learning (RL) or imitation learning (IL) often faces serious data scarcity issues~\citep{guiochet2017safety,zhan2021deepthermal}. To ensure reasonable performance, researchers often have to resort to laborious and costly data collection, tedious reset operation, and troublesome reward specifications~\citep{Zhu2020The}.
        Faced with restricted data acquisition, an alternative direction is to incorporate
        data generated with simulation or pre-collected from other domains (i.e. source domains) into policy learning for greater synergy~\citep{niu2022when,niu2023h2o+,vuong2023open}.
        However, source domains inevitably bear some domain gaps, such as appearance gaps~\citep{tobin2017domain}, dynamics gaps~\citep{peng2018sim}, and morphology gaps for embodied agents~\citep{gupta2021metamorph}.
        These domain gaps greatly restrict the usability of source domain data,
        as direct incorporation may negatively impact policy learning~\citep{niu2024comprehensive}.

        
        Existing cross-domain policy learning methods tend to design domain-specific policy transfer models equipped with domain correspondences~\citep{zhang2020learning}, corrections~\citep{eysenbach2020off}, or discriminations~\citep{stadie2016third,sharma2019third} that either rely on task-specific architecture designs (e.g. special encoders~\citep{mueller2018driving,wang2022versatile} and domain-specific regularizations~\citep{desai2020imitation,eysenbach2020off,niu2022when,xue2023state}), or are only applicable to a specific domain of data (e.g. adaptable vision encoder design only applicable to image inputs~\citep{rao2020rl}.
        Additionally, the domain-specific discriminators or mappings~\citep{stadie2016third,liu2018imitation} sometimes require tedious efforts to finetune for accommodating multiple source domains, substantially hindering data reuse efficiency and efficacy.
        The fundamental limitation of these approaches lies in their focus on bridging domain gaps within the policy learning process rather than addressing the root cause of the problem: the domain gaps in the data itself.
        While existing methods attempt to compensate for these gaps through complex model architectures and additional training objectives, they inevitably introduce new complexities and restrictions. This observation motivates us to rethink:
        \textbf{instead of adapting the policy learning process to accommodate domain gaps, can we directly bridge these gaps at the data level?}

        \begin{figure}
            \includegraphics[width=0.5\textwidth]{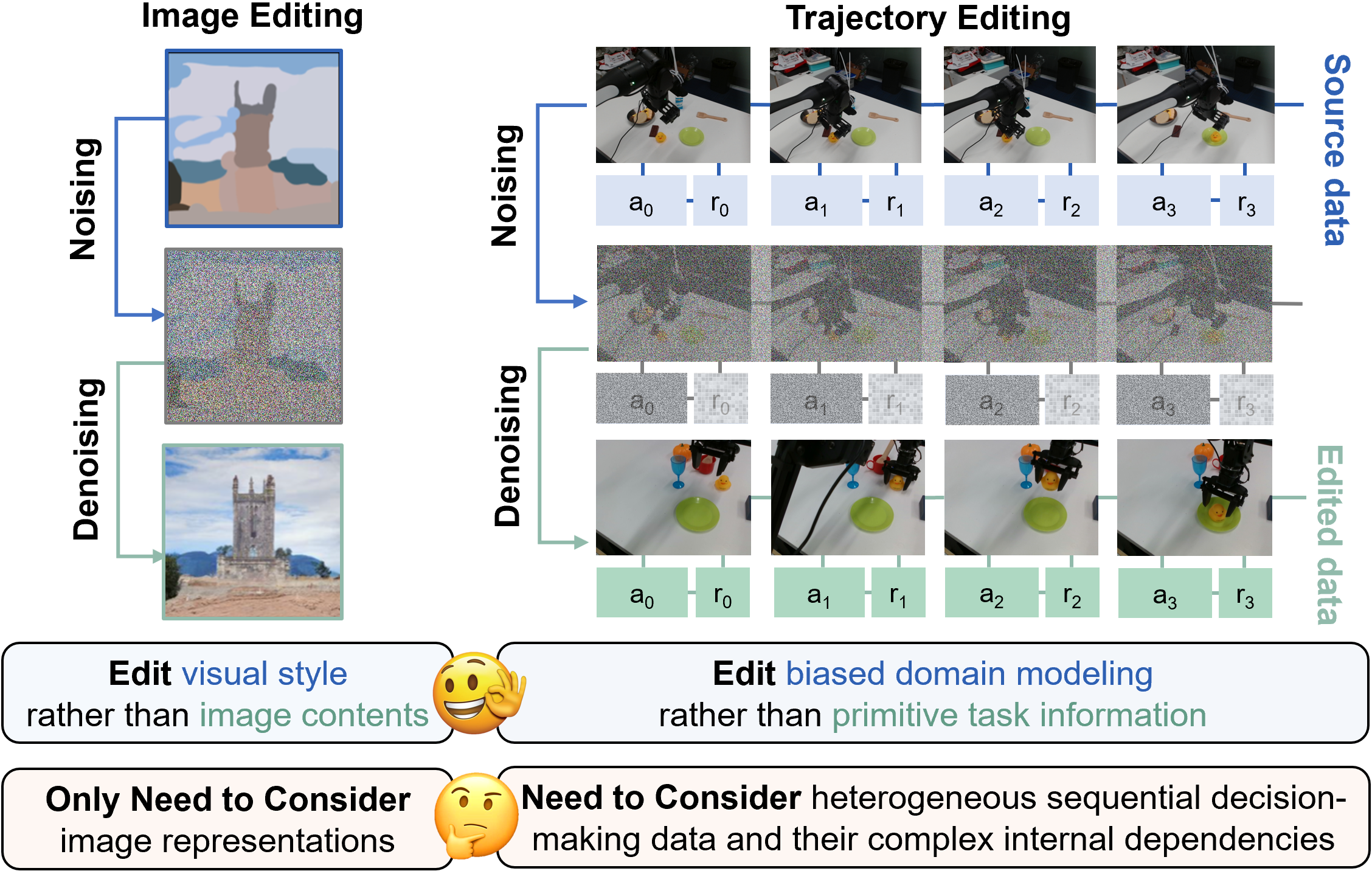}
            \caption{While sharing conceptual similarities with image editing, trajectory editing introduces distinct challenges due to the inherent complexity of sequential decision-making data, such as heterogeneous elements and complex internal dependencies.}
		      \label{fig:intro_analogy}
        \end{figure}


        Indeed, if source data can be directly transformed to minimize domain gaps with the target domain,
        we can flexibly choose any downstream policy learning method based primarily on task-specific considerations, bypassing the burden of cross-domain design complexities.
        However, this transformation requires a careful balance: we must modify the source data to align with target domain properties while preserving its essential primitive task information that can supplement policy learning from solely using target data.
        Traditional data augmentation and adaptation methods are insufficient for this purpose, since they typically overlook the benefits of the source data, and that is where ``editing'' becomes crucial.
        These pursuits naturally remind us of the diffusion-based image editing~\citep{meng2022sdedit,kawar2023imagic}, that transforms the visual style and aesthetic elements of an image without altering its narrative content, such as converting a stroke-painted image into a photorealistic one.
        However, as depicted in Fig.~\ref{fig:intro_analogy}, editing decision-making trajectory data from the source domain can be substantially more challenging, due to potential perception gaps and altered viewpoints in the state inputs, as well as the inherent dynamics gaps.
        Moreover, trajectory data in decision-making tasks are composed of fundamentally heterogeneous elements—such as observational states, actions, and rewards—that, if na\"ively treated as equivalent pixels in an image, would obscure the recognition of their respective inherence and delicate internal dependencies.
        Thus, despite being conceptually similar, image editing methods are not directly applicable to cross-domain trajectory editing due to the intrinsic heterogeneity and dependencies within trajectories and the complex domain discrepancies.

        To address these challenges, we propose the \textbf{Cross}-Domain \textbf{T}rajectory \textbf{ED}iting (\textbf{xTED}) paradigm with a novel diffusion model architecture tailored for decision-making data, which captures the target trajectory distribution as a prior.
        The proposed architecture features separate encoding and decoding of states, actions, and rewards to preserve their distinct physical meanings and internal temporal dependencies.
        To effectively capture the dynamics patterns within trajectories, we introduce dependency structure modeling mechanisms that capture the internal dynamics relationships among states, actions, and rewards.
        With the assistance of the above designs, we successfully extend the philosophy in diffusion-based image editing~\citep{huang2024diffusion} to decision-making data, equipping the xTED pipeline with three simple steps: 
        (1) train diffusion model on the target data; 
        (2) perturb source data with noises and then denoise them with the pre-trained diffusion model; 
        (3) incorporate edited source data into target data for policy learning with any algorithm at will.
        Through extensive experiments, we show that incorporating source data edited with xTED consistently yields performance gains over training solely on target data, while directly adding unprocessed source data often results in significant performance degradation, particularly in real-robot manipulation tasks.
        As a data adaptation method, xTED enables various choices of downstream single-domain policy learning methods to address cross-domain challenges, and can also be combined with other cross-domain methods to further enhance their performance.

	\section{Related Work}\label{rw}
	\subsection{Cross-Domain Policy Adaptation}
 Addressing the challenges posed by domain gaps has long been identified as an essential task for real-world policy learning~\citep{niu2024comprehensive}.
 The most straightforward approach is to construct direct mappings for state and action space between source and target domains~\citep{liu2018imitation,kim2020domain, zhang2020learning, raychaudhuri2021cross, wang2022weakly}.
 An alternative is to learn domain-agnostic task-relevant embedding with mutual information criterion~\citep{franzmeyer2022learn} and explicit domain discrimination~\citep{stadie2016third,sharma2019third}. It can also be achieved with temporal contrastive learning methods~\citep{Sermanet2017TCN,dwibedi2019temporal,yang2023polybot,choi2023efficient,li2024decisionnce} to learn representations that dependent on task progression while neglecting domain-variant information.
 Other than representation learning, it is also preferable to directly regularize the policy learning process under mild observation gaps, by reward augmentation~\citep{eysenbach2020off, liu2021dara, xue2023state} or reweighting the value update~\citep{niu2022when, niu2023h2o+, xu2023crossdomain}.
 All existing works suffer from complicated domain/task-specific designs, which lack flexibility and hinder convenient reuse and fine-tuning of those models to accommodate data from multiple source domains.
 Until now, there is no generic and flexible approach in the literature that handles domain gaps at the data level, and could potentially bypass the aforementioned drawbacks.
    \subsection{Diffusion Models for Decision Making}
    
    Recently, diffusion models have been adopted in various decision-making problem settings, including generating multi-modal policies~\citep{chi2023diffusionpolicy,wang2023diffusion}, single-step transitions~\citep{lu2023synthetic}, subgoals~\citep{black2024zeroshot}, trajectories~\citep{janner2022planning,ajay2023is,he2023diffusion,carvalho2023motion,luo2024potential}, and videos for planning~\citep{ajay2023compositional,yang2024learning,du2023learning,du2024video}. However, prior works primarily focus on single-domain generation tasks, such as data augmentation, rather than addressing cross-domain challenges.
    In contrast, xTED introduces a distinct setting by employing diffusion models to transform source trajectories to more closely resemble target domain characteristics while preserving useful primitive task information from source domains. 
    This setting is less data-intensive,
    as it does not require extensive training data as needed in diffusion-based generation tasks that aim to generate data from pure Gaussian noises.

    \section{Preliminaries}\label{headings}
    \noindent\textbf{Notations.}\quad
    We introduce the relevant notations and elements of decision trajectories with the standard formulation of Markov Decision Process (MDP).
    A finite-horizon MDP is defined as a tuple $\mathcal{M}:=\langle\mathcal{S}, \mathcal{A}, T, R, H, \gamma\rangle$ where $\mathcal{S}$ and $\mathcal{A}$ are state and action spaces, $T: \mathcal{S}\times\mathcal{A}\rightarrow\Delta_\mathcal{S}$ is the transition dynamics; $R: \mathcal{S}\times\mathcal{A}\rightarrow\Delta_\mathbb{R}$ is the reward function; $H$ is the trajectory horizon;
    $\gamma$ is the discount factor. The entire decision trajectory can be denoted with a sequence of transition tuples $\tau=\{(s_t, a_t, r_t)\}_{t=0}^{H-1}$, where $(s_t, a_t)\in\mathcal{S}\times\mathcal{A}$, $r_t\sim R(s_t,a_t)$ and $s_{t+1}\sim T(s_t,a_t)$.

    In this work, we focus on the cross-domain setting, where we could have one or many source domains, which differ from the target domain in terms of state (visual/viewpoint difference) and reward spaces as well as transition dynamics. Our goal is to adapt these source domain data to align with the target domain where preserving their primitive task information to facilitate target domain policy learning.

\noindent\textbf{Diffusion Model.}\quad
Diffusion models have emerged as a powerful class of generative models~\citep{DDPM}, which typically involve two phases: the forward and the reverse processes.
The forward process models the gradual addition of noise to the data, transforming the data distribution $p(\mathbf{x}_0)$ into a noised distribution over a series of discrete time steps $K$. This process can be described by the following equation:
\begin{equation}
    q(\mathbf{x}_{k}|\mathbf{x}_{k-1})=\mathcal{N}(\mathbf{x}_k;\sqrt{\alpha_{k-1}} \mathbf{x}_{k-1}, (1-\alpha_{k-1}) \mathbf{I}),\ \mathbf{x}_0\sim q(\mathbf{x}_0), k\in [1,\cdots, K]
\end{equation}
where $\alpha_k$ are the variance schedules. Given $\mathbf{x}_0$, we could sample the results of forward process at any noise step $k$: 
\begin{equation}
    \mathbf{x}_k\sim \mathcal{N}(\mathbf{x}_k;\mu_k,\sigma_k^2\mathbf{I}):=\mathcal{N}(\mathbf{x}_k;\sqrt{\alpha_1\alpha_2\cdots\alpha_k}\mathbf{x}_0,(1-\alpha_1\alpha_2\cdots\alpha_k)\mathbf{I})\label{xk}
\end{equation}
The reverse process, on the other hand, involves learning to denoise the data, effectively reversing the forward process:
\begin{equation}
p_\theta(\mathbf{x}_{k-1} | \mathbf{x}_k) = \mathcal{N}(\mathbf{x}_{k-1}; \mu_{\theta}(\mathbf{x}_k, k), \Sigma_k)
\end{equation}
where $\mu_{\theta}(\mathbf{x}_k, k)$ is the learnable mean of reverse conditional distribution.
The training process involves optimizing the variational lower bound of the data likelihood, a process that encourages the model to accurately reconstruct the data from noise.
This can boil down to minimizing the difference of the forward posterior and reverse conditional distribution $D_{KL}(q(\mathbf{x}_{k-1} | \mathbf{x}_k, \textbf{x}_0)|p_\theta(\mathbf{x}_{k-1} | \mathbf{x}_k))$ across all denoising steps, where $q(\mathbf{x}_{k-1} | \mathbf{x}_k, \textbf{x}_0)=\mathcal{N}(\mathbf{x}_{k-1}; \mu_q(\mathbf{x}_k, \mathbf{x}_0), \Sigma_q)$.
As the variance is only dependent on pre-defined $\alpha$ coefficients, we can construct the variance of the approximate reverse distribution $\Sigma_k$ also as $\Sigma_q=\sigma_q^2\mathbf{I}$. Thus, $\mu_{\theta}(\mathbf{x}_k, k)$ can be optimized towards the forward process posterior mean $\mu_q(\mathbf{x}_k, \mathbf{x}_0)$ as:
\begin{equation}
\begin{aligned}
    &\arg\min_\theta D_{KL}(q(\mathbf{x}_{k-1} | \mathbf{x}_k, \textbf{x}_0)|p_\theta(\mathbf{x}_{k-1} | \mathbf{x}_k)) \\&\Leftrightarrow \arg\min \frac{1}{2\sigma_q^2}\|\mu_\theta(\mathbf{x}_k,k)-\mu_q(\mathbf{x}_k, \mathbf{x}_0)\|_2^2
\end{aligned}
\end{equation}
In practice, we can adopt a further simplified surrogate loss instead of the distribution mean matching~\citep{DDPM}:
\begin{equation}
    \mathcal{L}_\theta=\mathbb{E}_{k\in[1,\cdots,K],\mathbf{x}_0\sim q(\mathbf{x}_0),\epsilon\sim\mathcal{N}(\mathbf{0},\mathbf{I})}\|\epsilon-\epsilon_\theta(\mathbf{x}_k,k)\|^2\label{eq:denoise}
\end{equation}
where $\epsilon_\theta$ represents a learnable network predicting the noise based on $\mathbf{x}_k$ and $k$, which is equivalent to estimating the forward posterior mean through re-parameterization trick.

\noindent\textbf{Diffusion-based image editing.}\quad
Image editing methods aim to modify visual styles while preserving the semantic content in source images, striking a balance between style realism in the target domain and content faithfulness to the source image. Unlike standard diffusion generative models that denoise fully noised inputs, diffusion editing methods~\citep{meng2022sdedit} observe that this desired partial modifications can be achieved by initiating the reverse process from an intermediate noise step $k<K$ of forward process as in Eq.~(\ref{xk}). This approach allows for selective adjustments in the source image, akin to how we want to transform source trajectories: removing domain biases while retaining valuable task-relevant information.
However, directly applying diffusion-based editing to decision-making data faces some technical barriers: 1) trajectories contain heterogeneous elements (states/actions/rewards) with distinct physical meanings, and 2) intricate dependencies exist among those elements. These challenges demand specially designed diffusion model architectures and tailored editing mechanisms as compared to editing in the image domain.

\section{xTED: Cross-Domain Trajectory Editing}\label{method}
 \begin{figure*}[t]
		\centering
		\includegraphics[width=\textwidth]{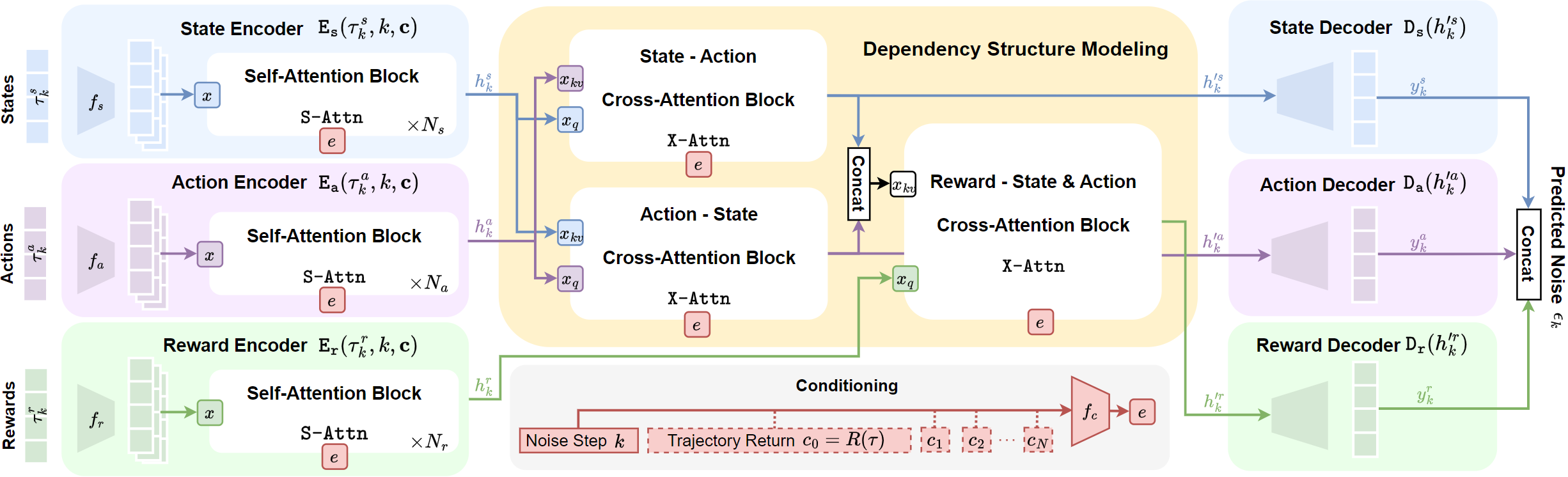}
		\caption{The model architecture is designed for capturing heterogeneous physical meanings of decision-making elements (states, actions, and rewards) and their intricate temporal and internal dependencies.}
        \label{fig:intro}
	\end{figure*}
In this section, we describe the detailed diffusion model architecture of xTED and its design philosophy to handle the intricate heterogeneity, sequential characteristics, and internal dependencies within the trajectory data, as illustrated in Fig.~\ref{fig:intro}. 
Next, we describe how to use it to effectively scaffold the cross-domain trajectory editing paradigm.

\subsection{Model Architecture}\label{ddit}
\noindent\textbf{Encoding and decoding design for trajectory data.}\quad
States, actions, and rewards in MDP present fundamentally distinct notions and physical meanings.
This naturally distinguishes the trajectory-based models from image-based models, which only consider spatial representations among visual patches~\citep{peebles2023scalable} and simply treat every pixel homogeneously in modeling designs~\citep{DDPM}.
Thus, xTED encodes and decodes state, action, and reward sequences $\tau^s,\tau^a,\tau^r$ in trajectories $\tau$ separately, which offers great flexibility to incorporate task-specific prior knowledge for each element.
In every noise step $k$, each sequence is first encoded into latent representation $h_k^{s}, h_k^{a}, h_k^{r}$ with 
using separate sub-networks $f$, and then processed with a self-attention block $\texttt{S-Attn}$ for temporal modeling along the sequence dimension (conditions are also embedded with $\mathbf{e}_k^c=f_c(k,\mathbf{c})$, short for $\mathbf{e}$ in Fig.~\ref{fig:intro}): 
\begin{equation}
        h_k^i=\texttt{E}_i(\tau_k^i,k,c):=\texttt{S-Attn}(f_i(\tau_k^i),\mathbf{e}_k^c), i\in[s,a,r]
\end{equation}
At the end, given the final representation $h_k^{\prime s}, h_k^{\prime a}, h_k^{\prime r}$, we use separate decoders to predict the corresponding noise $y_k$ at noise step $k$, which recover the original data dimension:
\begin{gather*}
    y_k^s=\texttt{D}_\texttt{s}(h_k^{\prime s})\in\mathbb{R}^{H\times|\mathcal{S}|},\ y_k^a=\texttt{D}_\texttt{a}(h_k^{\prime s})\in\mathbb{R}^{H\times|\mathcal{A}|},\
    y_k^r=\texttt{D}_\texttt{r}(h_k^{\prime r})\in\mathbb{R}^{H\times 1}
\end{gather*}
Crucially, this separation on encoding and decoding preserves the inherent differences between states, actions, and rewards, aiding the model in recognizing clear temporal dependencies on the latent representations $h_k$.
Compared with prior works that concatenate and process the state, action, and reward as an image-like big matrix~\citep{janner2022planning,he2023diffusion}, our design avoids exploiting spurious correlations within decision-making data and sidesteps modeling the joint trajectory distribution directly, thereby enhancing sample efficiency and maximizing model expressiveness.

\textbf{Dependency structure modeling.}\quad
Despite the distinction across states, actions, and rewards, significant internal dependencies exist. 
Specifically, unlike the 2D representation of images, the transition dynamics and reward function pose strong coupling and intricate internal dependency within sequential decision-making data.
To this end, the state and action embeddings \( h_k^s \) and \( h_k^a \) are cross-attended to capture mutual dependencies between states and actions by exchanging key-value pairs in multi-headed attention:
\begin{equation}
    h_k^{\prime s} = \texttt{X-Attn}(h_k^s, h_k^a,\mathbf{e}_k^c), \ h_k^{\prime a} = \texttt{X-Attn}(h_k^a, h_k^s,\mathbf{e}_k^c)
\end{equation}
Also, it is important to note that rewards are naturally dependent on state-action pairs, while this dependency does not work in the reverse direction due to their underlying causal relationships. 
Therefore, we query the reward embeddings \( h_k^r \) using concatenated state-action embeddings, without imposing unreasonable reward dependencies on state-action sequence modeling:
\begin{equation}
    h_k^{\prime r} = \texttt{X-Attn}(h_k^r, [h_k^s, h_k^a],\mathbf{e}_k^c)
\end{equation}
Introducing prior knowledge of the causal dependencies among states, actions, and rewards facilitates accurately capturing essential dependencies and temporal consistency within trajectory data, therefore improving sample efficiency and providing necessary regularizations when modeling complex trajectory distributions.

\textbf{Incorporating external conditions.} \quad
In addition to noise step \( k \), our model architecture accommodates various conditioning information \( \mathbf{c} = [c_i]_{i=0}^N \), guiding trajectories toward desirable patterns. 
For example, we can steer the entire trajectory modeling toward high-rewarding regions by conditioning on normalized trajectory returns \( \mathbf{c} = [R(\tau)] \)~\citep{ajay2023is}, with implementation details outlined in Appendix~\ref{app:imp_details}. 
Moreover, our model demonstrates superior synergy with return-conditioned modeling compared to na\"ive architectures that concatenate states, actions, and rewards along temporal or feature dimensions as input, as analyzed in Appendix~\ref{app:con_con}. This advantage arises from our separate encoding-decoding design and dependency structure modeling, which effectively recognize the causal dependencies between reward and state-action sequences.

\subsection{Cross-Domain Trajectory Editing with Diffusion}\label{xted}
\noindent\textbf{Training diffusion model on target data.}\quad
Given a target domain dataset, we first train our proposed diffusion model architecture in Fig.~\ref{fig:intro} to capture its trajectory distribution using diffusion loss Eq.~(\ref{eq:denoise}), where \(\mathbf{x}_k\) represents the noised trajectory \(\tau_k\). Additional conditions $\mathbf{c}$, if any, are concatenated with the noise step \(k\) as inputs.
Particularly, we keep the initial transition unchanged in each $H$-horizon trajectory during training, serving as an anchor for enhanced stability in long-sequence modeling.

\noindent\textbf{Editing source trajectories.}
The editing process contains two stages: the forward process that perturbs source data with noise and the reverse process that aligns noised data with target properties.

In the forward process, we obtain noised source trajectories \(\Tilde{\tau}_k\) from \(\mathcal{N}(\Tilde{\tau}_k; \Tilde{\tau}_0, \sigma^2_k \mathbf{I})\), which retains some information from the original source trajectories \(\Tilde{\tau}_0\). This procedure is governed by a ratio parameter \(\kappa = \frac{k}{K} \in [0, 1]\), determining the level of noise addition. A small \(\kappa\) preserves much of the source trajectory information, while a large \(\kappa\) diminishes it. 
The ideal selection principle for \(\kappa\) is to blur only fine-grained data details, such as transition dynamics in source data, while retaining valuable mesoscopic information, such as skill primitives in source trajectories. 
This noise initialization can yield favorable results by solving the reverse stochastic differential equation, which has achieved past success in diffusion-based image editing~\citep{meng2022sdedit,kawar2023imagic,ceylan2023pix2video}. 

In the reverse process, we denoise the noised \(\Tilde{\tau}_k\) using the learned target diffusion prior, aligning the edited trajectories \(\hat{\tau}_0\) with the target domain. 
The initial state-action-reward triple is kept unchanged during the editing process, consistent with the training process, which is crucial for enhancing editing performance as discussed in Appendix~\ref{app:con_con}.
Our empirical tests (see Appendix~\ref{app:t_ablation}) observe that setting \(\kappa = 0.5\) can already provide good enough results for incorporating the edited source data into policy learning, which ensures sufficient exploitation of source data while shifting its domain properties toward the target domain. 

\section{Experiments}\label{exp}

\subsection{Experimental Environment Setups}\label{setup}

\begin{figure*}[t]
    \centering
    \begin{subfigure}{\textwidth}  
        \centering
        \includegraphics[width=\textwidth]{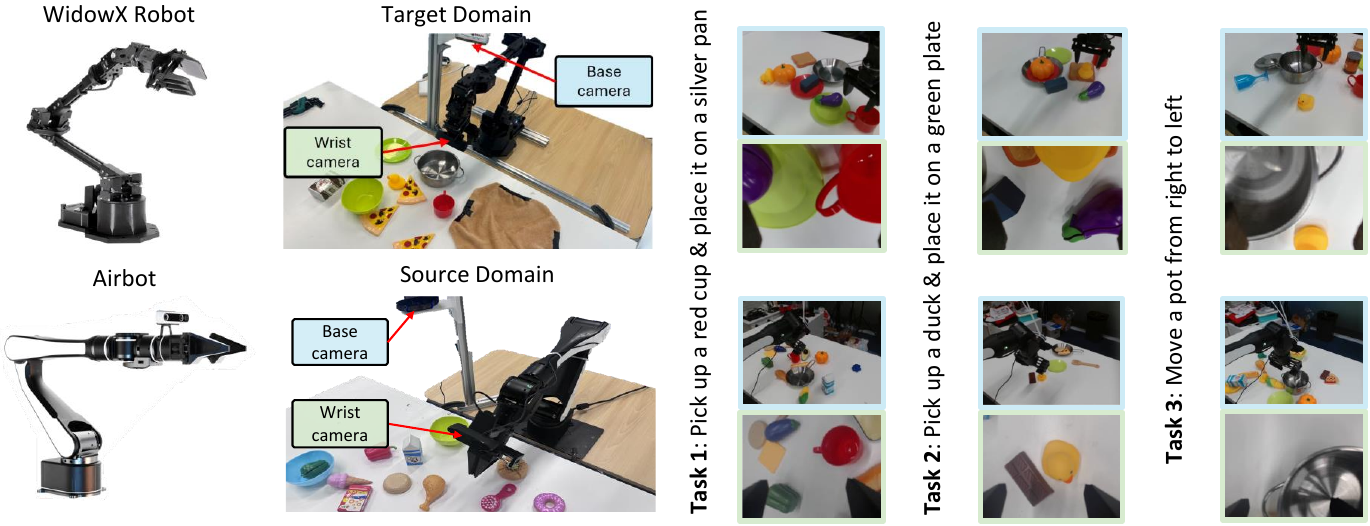}
    \end{subfigure}
    \vskip 4pt
    \begin{subfigure}{\textwidth}
        \centering
        \includegraphics[width=\textwidth]{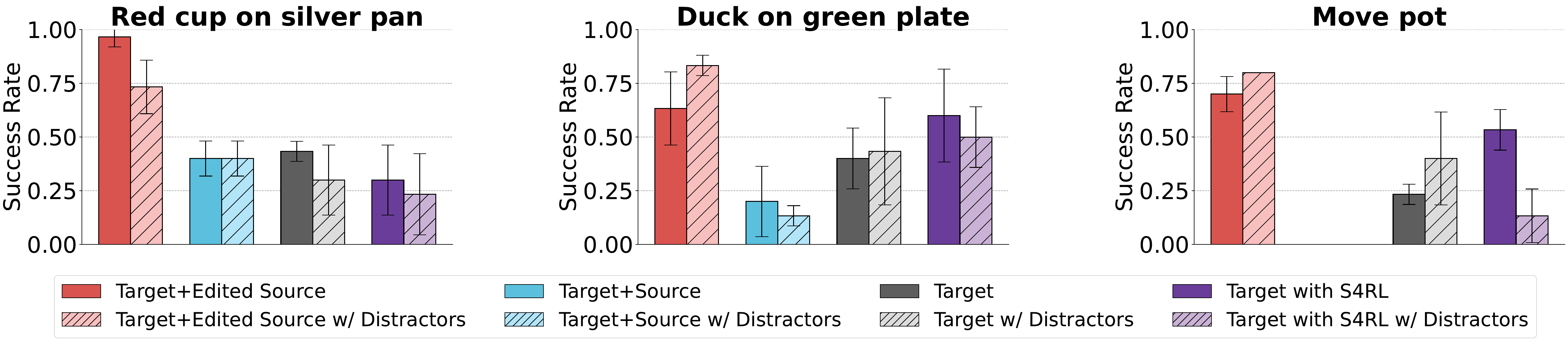}
    \end{subfigure}
    \caption{Target and source domains with complicated discrepancies on embodiments and viewpoints (top) and experiment results (bottom). 
    The top right presents the snapshots from base and wrist camera views of data collection processes in target/source domain from \textbf{Cup}/\textbf{Duck}/\textbf{Pot} tasks respectively. 
    The average success rate for real-robot tasks with/without distractors is obtained over 3 seeds.
    }
    \label{fig:embodi_gap}
\end{figure*}
	
 \noindent\textbf{Real-robot experiments.}\quad
We conduct real-world experiments in robotic environments where target data is collected by the WidowX robot and source data is collected by the Airbot, for 100 trajectories respectively. 
We build three manipulation tasks: (1) \textbf{Pick up a red cup and place it on a silver pan (Cup)}; (2) \textbf{Pick up a duck and place it on a green plate (Duck)}; (3) \textbf{Move a pot from right to left (Pot)}.
As shown in Fig.~\ref{fig:embodi_gap}, there are huge domain discrepancies on robot embodiments and camera viewpoints: two 7-DoF robots possess distinctive module shapes/lengths, embodiment masses and joint types; source and target configurations present different viewpoints and field of views from base and wrist cameras on two robots. 
Our objective is to incorporate the edited data from the Airbot to enhance policy performance on the WidowX robot.
 These task environments are highly stochastic, featuring randomly initialized object positions and poses, as well as distractors with various shapes, colors, and locations.
 For implementation, we edit trajectories with latent observation encoded with DecisionNCE~\citep{li2024decisionnce} and remove the reward-relevant modules in our model archiecture to fit the non-reward setting.
 Please see Appendix~\ref{app:robot_imp_detail} for more implementation details.

\textbf{Simulation experiments.}\quad
	We conduct simulation-based experiments using the MuJoCo physics simulator~\citep{todorov2012mujoco}. Specifically, we construct two source domains on Walker2d-v2 (WK) and HalfCheetah-v2 (HC), where we intentionally introduce dynamics and morphology gaps relative to the original target domains by modifying the physics modeling configurations:
	(1) \textbf{Gravity}: applying $2\times$ gravitational acceleration in the simulation dynamics;
	(2) \textbf{Friction}: using $0.25/0.5\times$ friction coefficient to make the agent harder to maintain balance; 
        (3) \textbf{Thigh Size}: using $2\times$ thigh size to introduce morphology gaps on the embodiment.
        We record the average return of Soft Actor-Critic~\citep{haarnoja2018soft} policies trained in these source domains and provide a straightforward quantification of the impact of domain gaps on source policy learning in Appendix~\ref{app:how_large_gap}.


For data acquirement and implementation details: (i) \textbf{target data}: we randomly select 20k transitions from the corresponding datasets in the standard offline RL benchmark D4RL~\citep{fu2020d4rl}. 
 Specifically, we only consider the Medium (M), Medium Replay (MR) and Medium Expert (ME) datasets, as we hardly use a random or expert policy for system control in common real-world scenarios. 
 (ii) \textbf{source data}: we collect trajectories of 20k transitions from the source domain with a SAC policy trained in the same domain.
See Appendix~\ref{app:imp_details} for more implementation details.

    \subsection{Baselines}\label{baseline}
(1) \textbf{Target}: training policies on the pre-sampled D4RL data (20k transitions per task) for simulation experiments and pre-collected WidowX data ($\sim$5k transitions per task) for real-robot experiments;
(2) \textbf{Source}: training policies on data (20k transitions per task) pre-sampled with SAC policies for simulation experiments;
(3) \textbf{Target with S4RL}: training policies solely on target data while applying the traditional data augmentation technique S4RL~\citep{S4RL} that adds action noise to target data;
(4) \textbf{Target+Source}: training policies on both target data and unprocessed source data (20k transitions collected by SAC source policies for simulation experiments/around 6k transitions from human tele-operation on Airbot for real-robot experiments);
(5) \textbf{Target+Edited Source} : training policies on both target data and source data edited with our pre-trained model model.

    \begin{table}[t]
\begin{minipage}[t]{0.5\textwidth}
\centering
\setlength\tabcolsep{3pt}
\caption{Average normalized scores for MuJoCo tasks on 20k target/source data (5 random seeds). $\Delta$ denotes the performance gain/degradation by adding source data (edited or not) against training on target data.}
\small
\adjustbox{scale=0.8,center}{
\begin{tabular}{ccl|c|c|cc|cc}
\toprule
\multicolumn{2}{c}{Target Data} & Domain Gap & Src & Tgt & Tgt+Src & $\Delta$ & Tgt+Src (Edited) & $\Delta$ \\
\midrule
\multirow{9}{*}{\rotatebox{90}{Halfcheetah}} 
&\multirow{3}{*}{\rotatebox{90}{Med}} & Gravity & 8.4$\pm$6.3  & 39.5$\pm$2.4 & \textbf{41.1$\pm$1.3} & \textcolor{gggreen}{+4.1\%} & \textbf{40.6$\pm$2.0} & \textcolor{gggreen}{+2.8\%} \\
&& Friction & 14.4$\pm$3.6 & 39.5$\pm$2.4 & 27.9$\pm$7.2 & \textcolor{ggdarkred}{-29.4\%} & \textbf{41.2$\pm$1.6} & \textcolor{gggreen}{+4.3\%} \\
&& Thigh Size & -0.1$\pm$1.0 & 39.5$\pm$2.4 & 40.3$\pm$2.2 & \textcolor{gggreen}{+2.0\%} & \textbf{40.7$\pm$2.4} & \textcolor{gggreen}{+3.0\%} \\
\cmidrule(lr){2-9}
&\multirow{3}{*}{\rotatebox{90}{Med-R}} & Gravity & 8.4$\pm$6.3 & 26.2$\pm$3.5 & 29.5$\pm$2.9 & \textcolor{gggreen}{+12.6\%} & \textbf{31.3$\pm$2.9} & \textcolor{gggreen}{+19.5\%} \\
&& Friction & 14.4$\pm$3.6 & 26.2$\pm$3.5 & 23.2$\pm$4.0 & \textcolor{ggdarkred}{-11.5\%} & \textbf{31.8$\pm$3.1} & \textcolor{gggreen}{+21.4\%} \\
&& Thigh Size & -0.1$\pm$1.0 & 26.2$\pm$3.5 & 29.0$\pm$3.7 & \textcolor{gggreen}{+10.7\%} & \textbf{33.0$\pm$3.0} & \textcolor{gggreen}{+26.0\%} \\
\cmidrule(lr){2-9}
&\multirow{3}{*}{\rotatebox{90}{Med-E}} & Gravity & 8.4$\pm$6.3 & 40.1$\pm$2.9 & 40.4$\pm$3.4 & \textcolor{gggreen}{+0.7\%} & \textbf{43.8$\pm$3.6} & \textcolor{gggreen}{+9.2\%} \\
&& Friction & 14.4$\pm$3.6 & 40.1$\pm$2.9 & 27.1$\pm$5.4 & \textcolor{ggdarkred}{-32.4\%} & \textbf{43.2$\pm$3.0} & \textcolor{gggreen}{+7.7\%} \\
&& Thigh Size & -0.1$\pm$1.0 & 40.1$\pm$2.9 & 37.4$\pm$5.5 & \textcolor{ggdarkred}{-6.7\%} & \textbf{43.0$\pm$3.0} & \textcolor{gggreen}{+7.2\%} \\
\midrule \midrule
\multirow{9}{*}{\rotatebox{90}{Walker2d}} 
&\multirow{3}{*}{\rotatebox{90}{Med}} & Gravity & 16.6$\pm$6.2 & 45.3$\pm$15.9 & \textbf{60.4$\pm$9.5} & \textcolor{gggreen}{+33.3\%} & \textbf{58.2$\pm$11.7} & \textcolor{gggreen}{+28.5\%} \\
&& Friction & 8.2$\pm$1.4 & 45.3$\pm$15.9 & 47.0$\pm$12.8 & \textcolor{gggreen}{+3.8\%} & \textbf{54.5$\pm$13.7} & \textcolor{gggreen}{+20.3\%} \\
&& Thigh Size & 7.6$\pm$3.5 & 45.3$\pm$15.9 & 49.6$\pm$12.2 & \textcolor{gggreen}{+9.5\%} & \textbf{58.9$\pm$11.7} & \textcolor{gggreen}{+30.0\%} \\
\cmidrule(lr){2-9}
&\multirow{3}{*}{\rotatebox{90}{Med-R}} & Gravity & 16.6$\pm$6.2 & 16.6$\pm$5.9 & 19.5$\pm$10.7 & \textcolor{gggreen}{+17.5\%} & \textbf{23.3$\pm$9.1} & \textcolor{gggreen}{+40.3\%} \\
&& Friction & 8.2$\pm$1.4 & 16.6$\pm$5.9 & 17.4$\pm$6.2 & \textcolor{gggreen}{+4.8\%} & \textbf{25.9$\pm$9.1} & \textcolor{gggreen}{+56.0\%} \\
&& Thigh Size & 7.6$\pm$3.5 & 16.6$\pm$5.9 & 18.0$\pm$6.7 & \textcolor{gggreen}{+8.4\%} & \textbf{25.9$\pm$9.1} & \textcolor{gggreen}{+56.0\%} \\
\cmidrule(lr){2-9}
&\multirow{3}{*}{\rotatebox{90}{Med-E}} & Gravity & 16.6$\pm$6.2 & 71.0$\pm$21.0 & 67.4$\pm$11.1 & \textcolor{ggdarkred}{-16.9\%} & \textbf{82.9$\pm$18.1} & \textcolor{gggreen}{+16.8\%} \\
&& Friction & 8.2$\pm$1.4 & 71.0$\pm$21.0 & \textbf{75.1$\pm$18.3} & \textcolor{gggreen}{+5.8\%} & \textbf{74.0$\pm$24.5} & \textcolor{gggreen}{+4.2\%} \\
&& Thigh Size & 7.6$\pm$3.5 & 71.0$\pm$21.0 & 77.0$\pm$19.3 & \textcolor{gggreen}{+8.5\%} & \textbf{81.0$\pm$21.4} & \textcolor{gggreen}{+14.1\%} \\
\midrule
&& Total & 165.3 & 716.1 & 727.3 & \textcolor{gggreen}{+1.6\%} & \textbf{833.2} & \textcolor{gggreen}{+16.4\%} \\
\bottomrule
\end{tabular}
}
\label{tab:main_exp}
\end{minipage}
\hfill
\begin{minipage}[t]{.5\textwidth}
		\centering
        \setlength\tabcolsep{3pt}
		\caption{Validation on xTED combined with cross-domain policy learning method DARA. Average normalized scores for MuJoCo tasks on 20k target/source data (5 random seeds).}
		\small
		\begin{tabular}{ccl|c|c}
			\toprule
			\multicolumn{2}{c}{Target Data} & Domain Gap & DARA & DARA+xTED \\
            \midrule
            \multirow{9}{*}{\rotatebox{90}{Halfcheetah}} 
            &\multirow{3}{*}{\rotatebox{90}{Med}} 
            & Gravity &  40.6$\pm$1.9 & \textbf{41.1$\pm$2.0} \\
            && Friction &  30.8$\pm$6.3 & \textbf{41.4$\pm$1.5} \\
            && Thigh Size &  40.4$\pm$2.4 & \textbf{41.5$\pm$1.6} \\
            \cmidrule(lr){2-5}
            &\multirow{3}{*}{\rotatebox{90}{Med-R}} 
            & Gravity &  29.9$\pm$3.0 & \textbf{31.1$\pm$3.1} \\
            && Friction &  22.8$\pm$3.7 & \textbf{32.3$\pm$3.1} \\
            && Thigh Size &  29.3$\pm$3.3 & \textbf{32.8$\pm$3.1} \\
            \cmidrule(lr){2-5}
            &\multirow{3}{*}{\rotatebox{90}{Med-E}} 
            & Gravity & 41.5$\pm$3.5 & \textbf{44.2$\pm$3.3} \\
            && Friction & 29.8$\pm$4.3 & \textbf{43.0$\pm$3.5} \\
            && Thigh Size & 37.9$\pm$5.2 & \textbf{43.3$\pm$3.4} \\
            \midrule\midrule
            \multirow{9}{*}{\rotatebox{90}{Walker2d}} 
            &\multirow{3}{*}{\rotatebox{90}{Med}} 
            & Gravity & 59.4$\pm$10.2 & \textbf{58.7$\pm$11.0} \\
            && Friction & 46.3$\pm$11.5 & \textbf{53.2$\pm$14.3} \\
            && Thigh Size & 48.5$\pm$11.6 & \textbf{56.8$\pm$12.9} \\
            \cmidrule(lr){2-5}
            &\multirow{3}{*}{\rotatebox{90}{Med-R}} 
            & Gravity & 20.3$\pm$9.6 & \textbf{22.5$\pm$8.9} \\
            && Friction & 17.9$\pm$5.7 & \textbf{24.7$\pm$8.5} \\
            && Thigh Size & 18.2$\pm$6.1 & \textbf{24.8$\pm$9.0} \\
            \cmidrule(lr){2-5}
            &\multirow{3}{*}{\rotatebox{90}{Med-E}} 
            & Gravity & 68.9$\pm$12.4 & \textbf{79.8$\pm$16.6} \\
            && Friction & 74.2$\pm$18.1 & \textbf{73.4$\pm$23.2} \\
            && Thigh Size & 75.3$\pm$17.3 & \textbf{78.6$\pm$19.4} \\
            \midrule
            && Total & 734.2 & \textbf{824.2} \\
			\bottomrule
		\end{tabular}\label{tab:dara}
\end{minipage}
\end{table}
        
\subsection{xTED with Various Policy Learning Methods}\label{sec:sim}
\textbf{xTED+single-domain IL.}\quad For real-robot experiments, we edit source data with xTED and then learn policies with behavior cloning (BC). The average success rates for \textbf{Cup}, \textbf{Duck}, and \textbf{Pot} tasks are shown in Fig.~\ref{fig:embodi_gap}.
Each result is estimated over 10 episodes per task with random environmental initializations and averaged over 3 seeds.
Target + Edited Source overwhelmingly outperforms the baselines in all the tasks, regardless of the presence of distractors.
Notably, in the \textbf{Cup} task, combining learning from edited source data increases the success rate from 43\% to 97\% and 30\% to 73.3\%, in environments without and with distractors respectively. 
In contrast, due to the huge domain gaps between the source and target domains, directly adding original source data yields no performance gain and sometimes even causes degradation against solely training on target data, e.g. 40\% to 20\% in \textbf{Duck} task and 23\% to 0\% in \textbf{Pot} task.
When evaluated with surrounding distractors, directly incorporating source data into policy learning sometimes leads to a more severe performance drop than without distractors, such as from 43\% to 13\% and from 40\% to 0\% in the \textbf{Duck} and \textbf{Pot} tasks, respectively. 
Furthermore, our method consistently outperforms the Target with S4RL across all tasks. This result highlights that data augmentation methods cannot reliably enhance policy performance when compared to using only the original target data. 
It underscores the superior effectiveness and necessity of the cross-domain trajectory editing paradigm, which appropriately leverages biased source data.
All numeric results in Fig.~\ref{fig:embodi_gap} can be found in Appendix~\ref{app:real_exp_tab} Table~\ref{tab:real_exp}.
    
\textbf{xTED+single-domain RL.}\quad In Table~\ref{tab:main_exp}, we present the comparative results of xTED and other baselines on the simulation-based experiments. 
We integrate our trajectory editing process with one of the widely-used state-of-the-art (SOTA) offline RL algorithms, Implicit Q-Learning (IQL)~\citep{kostrikov2022offline}. Notably, our method achieves the best or on-par performance in almost all tasks (18 out of 18) compared to the baselines. Specifically, we find that policy learning directly augmented with original source data can have negative effects compared to solely learning from target data in 5 out of 18 tasks, while augmenting with edited source data consistently shows improvement in all of the tasks.
    In WK-MR and HC-MR, edited source data helps improve the performance by over 50\% and 20\% respectively. The overall 16.4\% improvement also shows consistent stability on different domain gaps and quality of target data.
    This likely results from the fact that biased dynamics in source trajectories can detrimentally affect policy learning, despite the increased data coverage and diversified behavior patterns introduced by these trajectories. Nevertheless, xTED effectively adjusts the trajectories to realistic dynamics and provides meaningful augmentation, rather than merely expanding the data distribution.
    xTED also demonstrates the ability to handle various gaps simultaneously using a single fixed diffusion model in Appendix~\ref{app:multi-source}.

\noindent\textbf{xTED+cross-domain methods.}\quad
xTED can be integrated with any cross-domain IL/RL method to further enhance performance, as xTED helps reduce domain gaps at data level and ease the burden of downstream cross-domain policy learning.
To evaluate this, we combine xTED with DARA~\citep{liu2021dara} that corrects rewards in source data with domain discriminators. Specifically, we implement DARA on offline target and source data edited by xTED. The results in Table~\ref{tab:dara} demonstrate that xTED consistently improves DARA's performance since xTED helps increase the overlap of source and target distribution which could facilitate domain discriminator training in DARA.
Similar to DARA, many cross-domain policy adaptation methods rely on careful balance of source and target data or/and sufficient data availability to ensure effective domain correspondence/discriminator learning. Applying xTED for data pre-processing could reduce domain heterogeneity in target and source data, relaxing these requirements and enabling more data-efficient and robust cross-domain adaptation.

\begin{figure*}[t]
        \centering
        \includegraphics[width=\textwidth]{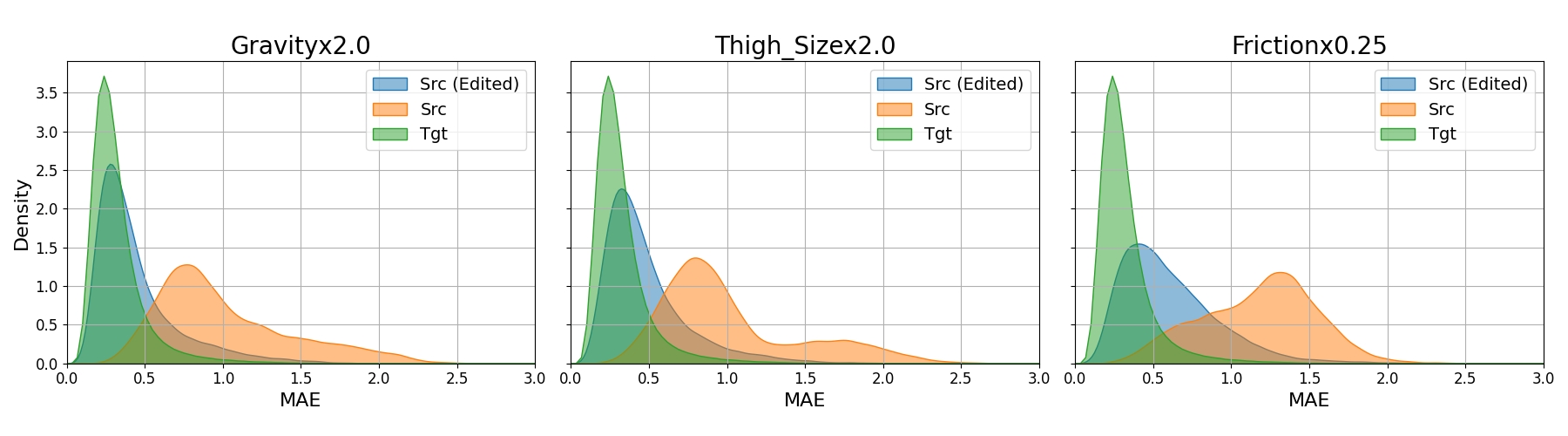}
        \caption{The distribution of dynamics errors (MAE) on source, edited source, and target data.}
        \label{fig:dynamics_error}
\end{figure*}

\noindent\textbf{Evaluation on editing quality.}\quad
To assess the effectiveness of dynamics adaptation achieved by our editing process, we learn a target-domain dynamics $p(s'|s,a)$ using MLP network with hidden sizes [2048, 1024, 512, 256] on HalfCheetah MR, ME, and M datasets (around 3.3M transitions) from D4RL, providing a reliable proxy for target dynamics in the HalfCheetah environment.
We then evaluate the mean absolute error (MAE) between the next states $\hat{s'}\sim p(\cdot|s,a)$ predicted by this target-domain dynamics model and the actual next states $s'$ from transitions $(s,a,s')$ sampled from three datasets: (i) the target dataset, (ii) the original source dataset, and (iii) the edited source dataset used in the HalfCheetah experiments shown in Table~\ref{tab:main_exp}. The results, visualized in Figure~\ref{fig:dynamics_error}, demonstrate that edited source data exhibits significantly lower dynamics errors than the original source data, reaching error levels comparable to those of the target data. This shows that edited data has good alignment with the target domain dynamics and thus can greatly complement the target dataset for policy learning. Mean square error distribution and all the numerical results are reported in Appendix~\ref{app:dynamics_error}.
Randomly selected original-edited source trajectory pairs are replayed in our \href{https://xted24.github.io/xTED/}{website}.

\subsection{xTED for Single-Domain Data Augmentation}\label{sec:augment}	
\begin{figure}[t]
    \begin{minipage}[t]{0.49\textwidth}
                \centering
                \includegraphics[width=\textwidth]{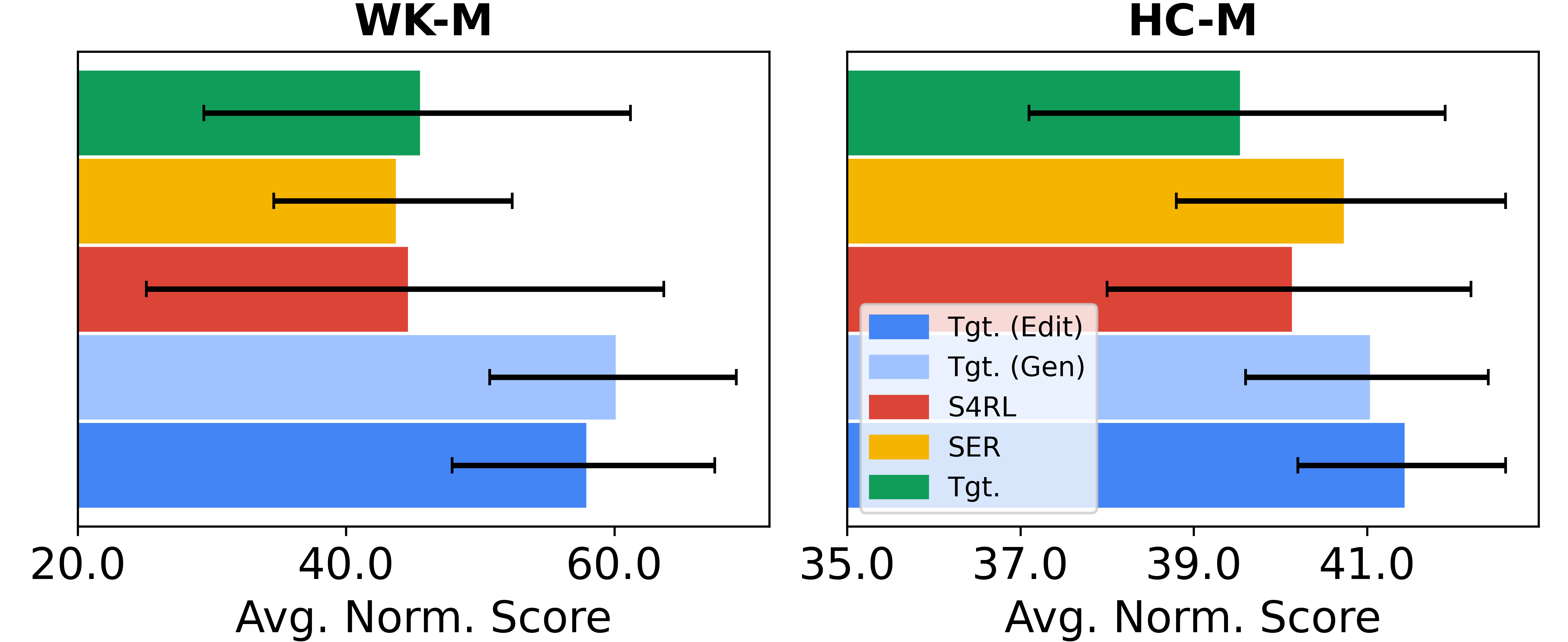}
                \caption{Average normalized scores for single-domain data augmentation over 5 random seeds.}
                \label{fig:data_aug}
    \end{minipage}%
    \hfill
    \begin{minipage}[t]{0.5\textwidth}
		\centering
            \setlength\tabcolsep{3pt}
		\captionof{table}{Architecture ablations (5 random seeds). \textcolor{ggdarkred}{Red} indicates lower scores than solely training on target data. }
		\small
		\adjustbox{scale=0.9,center}{
			\begin{tabular}{clccccc}
				\toprule
				Target Data & Domain Gap & TC & FC & Inv & CM & Ours
				\\
                \midrule
                    \multirow{3}{*}{HC-MR} & Gravity & \textcolor{ggdarkred}{23.8$\pm$3.2} & \textcolor{ggdarkred}{25.6$\pm$3.1} & 28.8$\pm$3.4 & \textcolor{ggdarkred}{20.1$\pm$3.2} & \textbf{31.3$\pm$2.9}\\
			& Friction & \textcolor{ggdarkred}{21.9$\pm$3.7} & \textcolor{ggdarkred}{22.4$\pm$4.0} & 27.9$\pm$3.6 & \textcolor{ggdarkred}{13.1$\pm$3.4} & \textbf{31.8$\pm$3.1} \\
                    & Thigh Size & 27.6$\pm$3.5 & 26.6$\pm$4.1 & 30.9$\pm$2.7 & \textcolor{ggdarkred}{19.3$\pm$2.1} & \textbf{33.0$\pm$3.0} \\
                    \midrule
				\multirow{3}{*}{WK-MR} & Gravity & {23.1$\pm$9.8} & \textcolor{ggdarkred}{15.9$\pm$6.9} & 20.8$\pm$6.3 & \textcolor{ggdarkred}{11.4$\pm$5.1} & \textbf{23.3$\pm$9.1}\\
				& Friction & {20.4$\pm$6.6} &{ 20.7$\pm$8.5 }& 20.7$\pm$7.6 & \textcolor{ggdarkred}{11.4$\pm$5.4} & \textbf{25.9$\pm$9.1} \\
                    & Thigh Size & 20.8$\pm$7.5 & {20.3$\pm$6.2} & 20.3$\pm$9.0 & \textcolor{ggdarkred}{9.6$\pm$4.0} & \textbf{25.9$\pm$9.1} \\
                    \midrule
                & Total & 137.6 & 131.5 & 149.4 & \textcolor{ggdarkred}{84.9} & \textbf{171.2}\\
                \bottomrule
		\end{tabular}}\label{tab:arch_ablation}
    \end{minipage}
\end{figure}

In reality, data from other domains is not always available. 
Fortunately, our model design also entails great potential to serve as a data generation model to navigate through this situation.
In Fig.~\ref{fig:data_aug}, we illustrate the results of using our model for data augmentation on small datasets (20k transitions) pre-sampled from D4RL.
Our model offers two simple types of augmentation approaches:
\textbf{(1) Tgt. (Edit)}: adding noise onto the original data as the forward process to match a theoretical Gaussian distribution ($\kappa=1.0$), followed by denoising with our model with the same number of steps. 
\textbf{(2) Tgt. (Gen)}: directly denoising standard Gaussian noise with our model. 
Model-based trajectory augmentation is compared with two baselines: \textbf{(1) S4RL}~\citep{S4RL}: adding a zero-mean noise distribution $\mathcal{N}(0,3e^{-4})$ on every dimension of the state space; 
\textbf{(2) SER}~\citep{lu2023synthetic}: learn a diffusion-based dynamics model for transition generation.
All methods augment the original dataset with the same number of synthetic samples.
It can be observed that augmentation with our diffusion model consistently leads in almost all tasks. 
S4RL achieves comparable scores in specific tasks but exhibits high variance due to its randomization mechanism. Compared with SER, our approach reveals that modeling longer-horizon trajectory dynamics and temporal consistency significantly improves augmentation quality for small datasets over merely modeling transition dynamics.


 






 \subsection{Ablations on Model Architecture}\label{ablation}
We ablate our model architecture to demonstrate the impact of different design insights and techniques.
We primarily consider four distinct architecture baselines:
\textbf{(1) Feature Concatenation (FC):} concatenating state, action, and reward sequences on feature dimension before passing through attention;
\textbf{(2) Transition concatenation (TC):} concatenating state, action, and reward sequences on temporal dimension 
before passing through attention;
\textbf{(3) Causal Mask (CM):} incorporating causal masks in attention blocks in our architecture for modeling Markovian process, considering transitions only depending on the previous step;
\textbf{(4) Diffusing over States + Inverse Dynamics (Inv):} solely diffusing over state trajectories, while employing two inverse models to obtain actions and rewards.
%

Table~\ref{tab:arch_ablation} demonstrates that our model consistently achieves superior performance, from which several key insights into architecture design can be derived:
(1) \textit{Entire trajectory modeling outperforms partial trajectory modeling}: 
Using inverse models (Inv) results in significantly higher compounding errors, as action feasibility and reward plausibility cannot be reliably maintained when only states are diffused without a unified model that handles all components consistently;
(2) \textit{Imposing Markovian properties as priors in trajectory modeling is suboptimal}: 
Incorporating Markovian modeling (CM) degrades the quality of edited source data, as local transition dynamics enforced by masked attention mechanisms may conflict with the global temporal consistency required in trajectory modeling. 
This suggests that trajectory modeling demands a distinct approach compared to common practices in decision-making approaches;
(3) \textit{Our design effectively captures intricate dependencies within trajectories}: 
Concatenating states, actions, and rewards along transition or feature dimensions (TC/FC) is problematic. It treats the entire trajectory as a homogeneous, image-like input, which obscures the model’s ability to correctly identify dependencies between transitions and decision elements, and introduces unwarranted dependencies of rewards on state-action sequences.
(4) \textit{Our design integrates well with return-conditioned diffusion}: 
Table~\ref{tab:condition} reveals that incorporating return conditioning in na\"ive architectures (TC/FC) leads to performance degradation, while in our model successfully guides edited source data towards higher-reward regions, resulting in superior policy performance. This is because only our design correctly models the one-way dependencies of reward and state-action sequences, enabling effectively implementing return-conditioned strategies.


     \begin{table}[t]
		\centering
        \setlength\tabcolsep{3pt}
		\caption{Our architecture is more compatible with (return-) conditioned diffusion. Results are averaged over 5 random seeds. \textcolor{ggdarkred}{Red} indicates lower scores than solely training on target data. $\uparrow$ indicates improved performance with return conditioning, while $\downarrow$ indicates the opposite.
}
		\small
		\adjustbox{scale=0.8,center}{
			\begin{tabular}{cl|cc|cc|cc}
				\toprule
				Target Data & Domain Gap & TC  & (Cond.) & FC & (Cond.) & Ours & (Cond.)
				\\
                \midrule
                    \multirow{3}{*}{HC-MR} & Gravity & \textcolor{ggdarkred}{23.8$\pm$3.2} & \textcolor{ggdarkred}{21.3$\pm$6.2} $\downarrow$ & \textcolor{ggdarkred}{25.6$\pm$3.1} & \textcolor{ggdarkred}{19.6$\pm$6.5} $\downarrow$ & \textbf{31.3$\pm$2.9} & 30.5$\pm$3.1 $\downarrow$\\
				& Friction & \textcolor{ggdarkred}{21.9$\pm$3.7} & \textcolor{ggdarkred}{23.8$\pm$5.6} $\uparrow$ & \textcolor{ggdarkred}{22.4$\pm$4.0} & \textcolor{ggdarkred}{20.1$\pm$7.9} $\downarrow$ & 31.8$\pm$3.1 & \textbf{35.1$\pm$1.8} $\uparrow$\\
                    & Thigh Size & 27.6$\pm$3.5 & \textcolor{ggdarkred}{21.1$\pm$7.7} $\downarrow$ & 26.6$\pm$4.1 & \textcolor{ggdarkred}{24.9$\pm$6.4} $\downarrow$ & 33.0$\pm$3.0 & \textbf{33.3$\pm$3.0} $\uparrow$\\
                    \midrule
				\multirow{3}{*}{WK-MR} & Gravity & 23.1$\pm$9.8 & \textcolor{ggdarkred}{8.9$\pm$5.2} $\downarrow$ & \textcolor{ggdarkred}{15.9$\pm$6.9} & \textcolor{ggdarkred}{7.5$\pm$4.2} $\downarrow$ & 23.3$\pm$9.1 & \textbf{24.5$\pm$9.0} $\uparrow$\\
				& Friction & {20.4$\pm$6.6} & \textcolor{ggdarkred}{9.0$\pm$5.5} $\downarrow$ & {20.7$\pm$8.5} & \textcolor{ggdarkred}{7.2$\pm$3.8} $\downarrow$ & \textbf{25.9$\pm$9.1} & 22.3$\pm$8.7 $\downarrow$ \\
                    & Thigh Size & 20.8$\pm$7.5 & \textcolor{ggdarkred}{11.9$\pm$6.3} $\downarrow$ & {20.3$\pm$6.2} & \textcolor{ggdarkred}{10.0$\pm$4.8} $\downarrow$ & 25.9$\pm$9.1 & \textbf{31.1$\pm$11.5} $\uparrow$\\
                    \midrule
                & Total & 137.6 & \textcolor{ggdarkred}{96.0} $\downarrow$ & 131.5 & \textcolor{ggdarkred}{89.3} $\downarrow$ & \textbf{171.2} & \textbf{176.8} $\uparrow$\\
                \bottomrule
		\end{tabular}}\label{tab:condition}
	\end{table}
    
\subsection{Other Results}
In the Appendix, we talk about the implementation details of simulation and real-robot experiments in App~\ref{app:imp_details}, as well as quantifying the magnitude of domain gaps introduced by simulation setups in App~\ref{app:how_large_gap}. Numeric results of real-robot experiments are listed in App~\ref{app:real_exp_tab}. More results on dynamics error evaluation are shown in App~\ref{app:dynamics_error}. We ablate different choices of editing ratio (\ref{app:t_ablation}), multiple editing iterations (\ref{app:editing_itr}), different amount of source data (\ref{app:data_ratio}), additional constraints and conditions on diffusion models (\ref{app:con_con}), different choices of trajectory horizons (\ref{app:horizon}), and different degrees of dynamics gaps (\ref{app:degree_gap}). We compare the performance and efficiency of Transformer and Temporal U-Net for diffusion-based data editing in App~\ref{app:tranUNet}. We also evaluate xTED's performance in situations with multiple source domains with different types of domain gaps in App~\ref{app:multi-source}. Moreover, the frequency analysis in App~\ref{app:freq} reveals that the domain gap between source and target trajectories is primarily concentrated in the medium-to-high frequency range, indicating that the philosophy of previous diffusion-based editing methods can also apply to decision-making trajectories.

\section{Conclusions and Limitations}\label{conclusion}
We present a generic, flexible, and effective trajectory editing paradigm (xTED), which reframes cross-domain policy adaptation as a data pre-processing problem. 
It is a task- and domain-agnostic method that avoids the need for task-specific policy adaptation designs and can accommodate multiple available source domains. 
Furthermore, it is compatible with any observation encoder and downstream policy learning method, making it adaptable to various task specifications.
Additionally, xTED is orthogonal to other cross-domain policy adaptation paradigms, which can be flexibly integrated for greater synergy if necessary. 
At the core of xTED, we introduce a scalable and flexible diffusion model architecture, effectively modeling the complex dependencies within trajectories. 
This design proves to be sample-efficient and highly compatible with (return-)conditioned diffusion. 
Besides, our model demonstrates strong potential as a data generator for single-domain data augmentation when no source data is available, highlighting its superiority in modeling entire trajectories.
Through extensive experiments, we show that incorporating source data edited with xTED consistently yields performance improvements over training solely on target data while directly adding unprocessed source data often results in significant performance degradation, particularly in real-robot manipulation tasks.
As for applicable scenarios, xTED can only handle situations where source and target domains have the same observation and action space structures.


\bibliography{sample}
\bibliographystyle{ACM-Reference-Format} 








\newpage
\appendix
\onecolumn
\section*{Appendix}
\section{Extended Discussions with Existing Works}\label{app:ext_diss}
xTED offers a generic, flexible, simple yet effective cross-domain trajectory editing paradigm that gains massive necessity and advantages over existing approaches. This section comprehensively elaborates on its distinctions from domain adaptation and data augmentation.

\subsection{Domain Adaptation}
xTED casts cross-domain policy adaptation as a data pre-processing problem, thereby eliminating the need for designing modality/task-specific adaptation modules that function during training. Most existing policy adaptation methods primarily focus on achieving \textbf{MDP Alignment}~\citep{kim2020domain}.

\paragraph{Existing research directions in domain adaptation.} The first category of approaches to achieving such alignment is through \textbf{learning direct mappings or corrections} between source and target domains. For observation space alignment, numerous works leverage advances in generative adversarial learning and contrastive learning to transform the observation space, achieving consistent visual rendering and camera viewpoints as in the target domain~\citep{zhu2017unpaired,liu2018imitation,zhang2019,bewley2019learning,rao2020rl,Liu2020State,Gangwani2020State-only,radosavovic2021state}.
In scenarios with dynamics mismatch, it becomes necessary to align both state and action spaces simultaneously, which introduces more complex correspondence learning mechanisms such as (dynamics/temporal) cycle consistency~\citep{kim2020domain,zhang2020learning,raychaudhuri2021cross,zakka2021xirl,wang2022weakly} and mutual information criteria~\citep{franzmeyer2022learn}.
Alternatively, modifying the reward space can compensate for the misalignment in transitional dynamics within MDPs~\citep{eysenbach2020off,liu2021dara,lyu2024crossdomain,xue2023state}, which may also be implicitly addressed by adaptively regularizing Q-values~\citep{niu2022when,niu2023h2o+,hou2024improving}.
\textit{This category necessitates model retraining when new data from other source domains becomes available, which can be time-consuming and labor-intensive.}

The second category resorts to \textbf{building a canonical representation space} that preserves task-relevant yet domain-agnostic information~\citep{stadie2016third,Pan2017b, mueller2018driving,Sermanet2017TCN,cetin2021domainrobust,wang2022versatile,yang2023polybot,choi2024domain}.
This auxiliary representation space, often viewed as skills~\citep{gupta2017learning,hejna2020hierarchically,pertsch2022cross,xu2023xskill} or subgoals~\citep{sharma2019third,ma2022vip,ma2023liv,li2024decisionnce}, facilitates the construction of aligned domain-invariant MDPs by capturing task semantics that effectively guide agents in completing tasks.
\textit{However, this approach involves designing task-specific loss functions and optimization procedures, which may struggle to find a representation space that effectively encapsulates the inputs and reflects the task.}

Despite the aforementioned shortcomings, these two categories share a common Achilles' heel: mode collapse~\citep{durall2021combating}. Most policy adaptation methods, whether explicitly or implicitly, rely on the high-level philosophy of generation and discrimination. 
\textit{Learning an informative mapping or a canonical representation that can effectively recover the entire target distribution can be exceedingly challenging, especially due to data imbalance and the sensitivity of hyper-parameter tuning.}

\paragraph{Why ``data adaptation''?} To overcome all the aforementioned issues, it is crucial to avoid designing modality/task-specific adaptation modules and instead \textbf{handle domain adaptation at the data level}. This approach provides significant flexibility, allowing the integration of various upstream observation encoders for different modalities, as well as diverse downstream policy learning methods that best suit the tasks at hand.

\paragraph{Why diffusion-based data editing?} Direct data transformation cannot rely on a pre-trained deterministic mapping or representation to pre-process source data, as this would face the same challenges as traditional domain adaptation methods, leading to transformed data with reduced variety and insufficient coverage in the target distribution. Moreover, such mappings or representations require re-training or fine-tuning to accommodate multiple source domains. In contrast, diffusion-based editing ensures diversity in the transformed data due to its stochastic nature and avoids mode collapse. Additionally, the editing process is universally domain-agnostic, correcting only the dynamics and observation patterns while preserving task-relevant information in source data with a target distribution prior modeled by a diffusion model, thus enabling a retraining-free approach.

Above all, xTED represents a \textbf{novel, generic, and flexible} cross-domain data-level policy adaptation paradigm. It is \textit{domain-agnostic}, accommodating multiple source domains without the need for re-training or fine-tuning the diffusion model; \textit{modality-agnostic}, being compatible with various observation encoders; \textit{task-agnostic}, integrating seamlessly with different task-specific policy learning methods; \textit{compatible} with other policy adaptation paradigms and approaches for greater synergy.

\subsection{Distinctions from Data Augmentation}
\paragraph{Setting.} Data augmentation is primarily employed in single-domain settings, which are fundamentally different from xTED's cross-domain applications. Typically, no data from other domains is introduced, and data augmentation focuses on manipulating training data to enhance robustness, e.g. noising the original data~\citep{S4RL}.

In contrast, data editing is a broader paradigm that encompasses the applications of data augmentation. Even when data from other domains is unavailable, xTED can serve as an expressive data generation model for augmenting small-sample datasets in single-domain settings, outperforming state-of-the-art data augmentation strategies~\citep{SER,S4RL} (see Section \ref{sec:augment}).

\paragraph{Diffusion-based data editing V.S. data augmentation.}We have to emphasize that using diffusion models for data editing is both sample-efficient and error-tolerant than diffusion-based trajectory generation for data augmentation~\citep{lu2023synthetic,he2023diffusion,jackson2024policy}. This is because the goal in data editing is far more easily attainable compared to its application in data generation tasks.
When diffusion models are used for data generation, the focus is usually on achieving a high level of precision to generate data that closely matches the original data distribution.
However, in the context of data editing, the requirements are relaxed to simply distort (edit) the data to more closely match the target dynamics and observation patterns than the source data. 
Crucially, there is no need for edited source data to closely resemble target data in every detail.
Thus, diffusion models for editing are relatively insensitive to the amount of target data used, having minimal impact on the ultimate policy performance.

\subsection{Applicability to Diverse Downstream Task Settings}
The xTED framework can serve as an extra data pre-processing step with great flexibility. The edited source data can be integrated with target data for downstream policy learning using any algorithms (IL or RL, single- or cross-domain).
Our proposed diffusion architecture is also compatible with multi-modal observations, including proprioceptive, exteroceptive (visual/point clouds), and language inputs, as long as modality-specific encoders are employed.
If these task-specific elements exhibit similarities with rewards, which have specific dependencies on states and/or actions, the architecture can be easily extended with minimal additional design complexities by incorporating extra encoders, decoders, and attention blocks, akin to our reward dependency design.
Furthermore, practitioners can also omit the reward-relevant modules to fit non-reward (imitation learning) settings, as demonstrated in our real robot experiments (Section \ref{sec:sim}).

 \section{Implementation Details and Experiment Setup}\label{sec:app_details}

	\subsection{Simulation Experiments}\label{app:imp_details}
	The implementation details in our simulation experiments are specified as follows:
	\begin{itemize}[leftmargin=*, topsep=0pt]
        \item \textbf{Model and Method Setups}: to strike the balance among the concerns from Separate Encoding-Decoding in Section~\ref{ddit}, we try to encode the state, action, and reward into $16\times$ embedding dimensions respectively. 
        In terms of the number of self-attention blocks, we set $N_s=N_a=2, N_r=1$ in Figure \ref{fig:intro} for the least computation consumption with performance guarantee.
        The training process takes denoising steps $K=200$ and optimizes over Eq.~\ref{eq:denoise}.
        In xTED, we perturb the source trajectories with $k=100$ steps of gaussian noises and denoise them with the pretrained model in $k=100$ steps ($\kappa=0.5$). 

 The computation of attention blocks is proportional to $9H^2$ (self-attention: $(N_s+N_a+N_r)^2$; cross-attention:$4H^2$), so we also set architecture with minimal differences on computation efforts for the architecture ablation baselines for fair comparisons, i.e. FC, TC, Inv, CM.
For CM, we keep the architecture the same with xTED, with the only difference of adding a causal mask with markovian constraints.
We use 9 successive self-attention blocks and FC ($9\times H^2$), while one self-attention block for TC ($(3\times H)^2$), achieving the same attention computation.
For Inv, we only have to diffuse over states without addressing the intricate dependencies within trajectories so we reduce the computation and adopt only 2 self-attention blocks for efficiency.
 
            \item \textbf{Data Usage}.
            We use only 20k transitions from the D4RL target dataset for diffusion training and policy learning on each task, which represents merely 1/50 of HC-M, 1/10 of HC-MR, 1/100 of HC-ME, 1/50 of WK-M, 1/15 of WK-MR, and 1/100 of WK-ME from the original D4RL dataset. The full dataset is already sufficient to train a high-performing policy, making it unnecessary to combine source data and risk potential negative effects due to domain gaps.

	    \item \textbf{Encoder-Decoder}. 
         Separately encoding and decoding state, action, and reward sequences in trajectories offers great flexibility to apply task-specific prior knowledge in constructing a reasonable information bottleneck.
        The spaces of embedded state $z_s$, action $z_a$, and reward $z_r$ should not be overly compressed as following attention blocks requires sufficiently rich information to discover connections.
        On the other hand, the embedding spaces should not be infinitely enlarged due to concerns on computational complexity and memory use.
        From an interactive angle, the sizes of these embedding spaces should not be diversely imbalanced so that the model is less prone to ignoring the low-dimensional action and reward embeddings when involved with visual observations.
        In all the transformer architectures ablated in Section~\ref{ablation}, the encoders and decoders we used are simple MLPs with two dense layers:
\begin{equation}
\begin{aligned}
    z_s &= \mathrm{MLP}(s) \\
    z_a &= \mathrm{MLP}(a) \\
    z_r &= \mathrm{MLP}(r) \\
    \mathbf{z} &= [z_s, z_a, z_r]
\end{aligned}
\end{equation}
        
        In xTED and CM, we expand the observation, action, and reward into a feature (hidden) space with 16 times the original dimensionality. 
        In Inv, FC, and TC, each observation, action, and reward is embedded into a 320-dimensional space.

	    \item \textbf{Self-Attention Block $\texttt{S-Attn}(x,c)$}.
        In xTED, TC, FC, Inv and CM, we use multi-head self-attention (MSA) mechanism to apply long-horizon consistency in sequence modeling. Specificlly, we use the \texttt{MultiHeadDotProductAttention} (MSA) function implemented by JAX, denoted by $\mathrm{ATTN}$. The forward processes of self-attention block are formulated as follows:
        \begin{equation}
        \begin{aligned}
            &z_c = \mathrm{MLP}(c) \\
            &q=k=v=\mathrm{LayerNorm}(x)\cdot(1+z_{c,\mathrm{scale}}) + z_{c,\mathrm{shift}}\\
            &x = x+z_{c,\mathrm{gate},\mathrm{attn}}\cdot\mathrm{ATTN}\left(q,k,v\right) \\
            &x = x+z_{c,\mathrm{gate},\mathrm{mlp}}\cdot\mathrm{MLP}\left(\mathrm{LayerNorm}(x)\cdot(1+z_{c,\mathrm{scale}}) + z_{c,\mathrm{shift}}\right)
        \end{aligned}
        \end{equation} 
        Here, $x$ is the input, $c$ is the condition, $z_{c,\mathrm{gate}}, z_{c,\mathrm{scale}}$ and $z_{c,\mathrm{shift}}$ are chunked from $z_c$.
        More detailed hyperparameters are shown in Table~\ref{tab:hyper}.

        \item \textbf{Cross-Attention Block $\texttt{X-Attn}(x_q,x_{kv},c)$}.
        In xTED and CM, we use cross-attention mechanism to apply sequence modeling. 
        The input argument \( c \) is produced by the output of the condition encoder \( f_c \), which splits into two components, \( c_q \) and \( c_{kv} \), ensuring that the tensor shapes align with \( x_q \) and \( x_{kv} \), respectively.
        We still use the \texttt{MultiHeadDotProductAttention} implemented by JAX. The forward processes of cross-attention blocks are formulated as follows:
        \begin{center}
            \begin{equation}
                \begin{aligned}
                &(c_q, c_{kv}) = c \\
                &z_{c_q} = \mathrm{MLP}(c_q) \\
                &z_{c_{kv}} = \mathrm{MLP}(c_{kv}) \\
                &q = \mathrm{LayerNorm}(x_q)\cdot(1+z_{c_q,\mathrm{scale}}) + z_{c_q,\mathrm{shift}} \\
                &k = v = \mathrm{LayerNorm}(x_{kv})\cdot(1+z_{c_{kv},\mathrm{scale}}) + z_{c_{kv},\mathrm{shift}} \\
                &x_{kv} = x_{kv}+z_{c_{kv},\mathrm{gate}, \mathrm{attn}}\cdot\mathrm{ATTN}(q,k,v) \\
                &x_{kv} = x_{kv}+z_{c_{kv},\mathrm{gate}, \mathrm{mlp}}\cdot\mathrm{MLP}\left(\mathrm{LayerNorm}(x_{kv})\cdot(1+z_{c_{kv},\mathrm{scale}}) + z_{c_{kv},\mathrm{shift}}\right)
                \end{aligned}
            \end{equation}
        \end{center}
        Here, $x_q$ is the input for query, $x_{kv}$ is the input for key and value, $c_q$ is the condition for $x_q$, $c_{kv}$ is the condition for $x_{kv}$.
        $z_{c_{kv},\mathrm{gate}}, z_{c_{kv},\mathrm{scale}}$ and $z_{c_{kv},\mathrm{shift}}$ are chunked from $z_{c_{kv}}$. $z_{c_q,\mathrm{scale}}$ and $z_{c_q,\mathrm{shift}}$ are chunked from $z_{c_{q}}$.
        More detailed hyperparameters are shown in Table~\ref{tab:hyper}.

        \item \textbf{Editing Details}.
        When we use xTED for trajectory editing, the step parameter $\kappa$ we use is 0.5, while the total time steps $K$ we used for training xTED model is 200. This implies that we add $k=100$ steps of noise to the original source-domain data, then denoise it with xTED with the same number of steps.

        \item \textbf{Policy Learning}.
        In this paper, we use IQL and TD3+BC as policy learning algorithms. The key hyperparameters we used are shown in Table~\ref{tab:hyper}, which are basically the same as the original papers.
	
        \item \textbf{Evaluation}: All the simulation results are obtained by averaging over the same set of 5 random seeds.
	
	\item \textbf{Computing Resources}: we ran experiments largely on 8 NVIDIA A100 GPUs via an internal cluster.
        \end{itemize}
	
	\begin{table}[t]
		\centering
		\caption{Hyperparameters for simulation (sim) and real-robot (real) experiments.}
			\begin{tabular}{lc}
				\toprule
				Hyper-parameter & Value\\
				\midrule
				\rowcolor{gray!20}\multicolumn{2}{c}{\textbf{xTED} (sim)} \\
                Number of self attention blocks for state & 2
                \\
                Number of self attention blocks for action & 2
                \\
                Number of self attention blocks for reward & 1
                \\
                Number of cross attention blocks for s-a & 1
                \\
                Number of cross attention blocks for a-s & 1
                \\
                Number of cross attention blocks for r-sa & 1
                \\
                Size of hidden dim for encoder/decoder & 16$\times$original dim 
                \\
                Trajectory horizon $H$ & 20
                \\
                \rowcolor{gray!20}\multicolumn{2}{c}{\textbf{xTED} (real)} \\
                Number of self attention blocks for state & 2
                \\
                Number of self attention blocks for action & 2
                \\
                Number of cross attention blocks for s-a & 1
                \\
                Number of cross attention blocks for a-s & 1
                \\
                Size of hidden dim for encoder/decoder & 1$\times$original dim 
                \\
                Trajectory horizon $H$ & 5
                \\
                \rowcolor{gray!20}\multicolumn{2}{c}{\textbf{IQL} (sim)} \\
                Actor learning rate & 3e-4
                \\
                Critic learning rate & 3e-4
                \\
                Optimizer & adam
                \\
                \rowcolor{gray!20}\multicolumn{2}{c}{\textbf{TD3+BC} (sim)} \\
                Actor learning rate & 1e-5
                \\
                Critic learning rate & 1e-3
                \\
                Optimizer & adam
                \\
                \rowcolor{gray!20}\multicolumn{2}{c}{\textbf{BC} (real)} \\
                Actor learning rate & 1e-4
                \\
                Optimizer & adam
                \\
				\bottomrule
		\end{tabular}\label{tab:hyper}
	\end{table}

 \subsection{Real-Robot Experiments}\label{app:robot_imp_detail}
In addition to configurations identical to those used in simulation experiments, we summarize below the specific implementation details for the real-robot experiments:
 	\begin{itemize}[leftmargin=*, topsep=0pt]
        \item \textbf{Model and Method Setups}: 
        Different from simulation experiments, the robot manipulation tasks involves environments where rewards are hard to specify, thus resulting in collecting non-reward trajectories.
        Correspondingly, the self-attention and cross-attention blocks for reward trajectories are removed from the original xTED architecture, leaving only self-attention and cross-attention blocks for state and action sequences. In terms of other architecture designs and editing details, we keep them identical with simulation experiment setups.
        We set the trajectory horizon for modeling with xTED to $H=5$, as the robot manipulation trajectories contain higher-dimensional features, necessitating a trade-off in horizon length to reduce computational burden.
        \item \textbf{Data Collection}.
        Each source/target dataset comprises 100 trajectories (40-60 transitions per trajectory with a frequency of around 5Hz, 75 trajectories w/ distractors and 25 trajectories w/o distractors) with RGB observations (480 × 640) from an external camera and a camera mounted on the robot’s gripper, as well as end-effector actions.
        \item \textbf{Observation Pre-processing}.
        We choose DecisionNCE-T~\citep{li2024decisionnce} to compress the 480x640 camera views into embedding space with 1024 feature dimensions before applying xTED. xTED is then deployed on the lower-dimensional observation representation instead of the high-dimensional image observations.
        \item \textbf{Encoder-Decoder.} In xTED architecture, we maintain the original feature dimensionality of the observation representation (1024) and expand the action space by two times orginal feature dimensionality.
        \item \textbf{Transformer Architecture.} The architecture remains identical to that used in our simulation experiments, with the only exception being the removal of modules related to reward dependency modeling.
        \item \textbf{Policy Learning}.
        The robot policy network consists of three layers of MLPs for action prediction, trained via vanilla behavior cloning (BC) with L2 regression using the Adam optimizer at a learning rate of 1e-4. All models are trained for 2.5k epochs with a batch size of 64.
        \item \textbf{Evaluation}.
        We evaluate the policy checkpoint with 10 episodes for each seed with random environmental initializations and 3 random seeds for each result.
	
	\item \textbf{Computing Resources}: we ran experiments on 4x NVIDIA A800 GPUs via an internal cluster.
        \end{itemize}

    \subsection{On the Magnitude of Domain Gaps Introduced by the Simulation Experiment Setups}
\label{app:how_large_gap}
    In Section~\ref{setup}, we mentioned that we modified the original dynamics—such as 2x gravity, 0.25/0.5x friction (HC/WK), and 2x thigh size—to construct source domains with biased dynamics. To quantify the domain gaps, we recorded the mean values of the average returns of SAC policies trained in the source domain but evaluated in the target domain, and compared them to SAC policies trained and evaluated in the target domain. As shown in Table~\ref{tab:quantify_gap}, the domain gaps significantly hinder direct policy transfer from the source to the target domain, resulting in notable performance degradation.

    \begin{table}[H]
        \centering
        \caption{\small Average scores of SAC policies trained in source/target domain and evaluated in target domain.}
        \begin{tabular}{llcc}
        \toprule
        \textbf{Task} & \textbf{Domain Gap} & \textbf{Avg. Return (Target)} & \textbf{Avg. Return (Source)} \\
        \midrule
        \multirow{3}{*}{Halfcheetah} & Gravity    & 4513 & \multirow{3}{*}{10290} \\ 
                                     & Friction   & 5099 &                        \\ 
                                     & Thigh Size  & 5537 &                        \\ 
        \midrule
        \multirow{3}{*}{Walker2d}    & Gravity    & 1233 & \multirow{3}{*}{4916}  \\ 
                                     & Friction   & 3392 &                        \\ 
                                     & Thigh Size  & 437  &                        \\ 
        \bottomrule
        \end{tabular}
        \vskip -0.1in
        \label{tab:quantify_gap}
    \end{table}

\section{Additional Experiment Results}\label{app:extra_ablations}
\subsection{Numeric Results of Real-Robot Experiments}\label{app:real_exp_tab}
The values of success rate depicted in Fig.~\ref{fig:embodi_gap} bar plots are listed in the following Table~\ref{tab:real_exp}.
Surprisingly, incorporating edited source trajectories helps improve success rate of task completion by a fairly large margin, i.e. $>100\%$ or even $>200\%$ against solely training on target data.
However, directly combining original source trajectories generally results in severe performance degradation, i.e. in 5/6 tasks.
Particularly, the success rates drop to 0\% in Pot tasks where the robot arm cannot identify the target object and approach it.

Qualitatively, as shown in the recorded videos on our \href{https://xted24.github.io/xTED/}{website}, robot policies trained on target data consistently exhibit poor performance in lifting the end effector, preventing the cup from being placed through the top opening of the pan. Additionally, these policies show low accuracy and poor timing in grasping or releasing target objects, attributed to the insufficiency and narrowly distributed target data. Conversely, directly incorporating source data results in policies that are more aggressive and less precise in control, as Airbot operates within a broader range of control dynamics.

\begin{table}[h!]
\centering
\caption{Performance comparisons across different real-robot tasks and baselines.}
\scalebox{0.9}{
\begin{tabular}{lcccc}
\toprule
\textbf{Task} & \textbf{Target+Edited Source (Ours)} & \textbf{Target+Source} & \textbf{Target} & \textbf{Target (w/ S4RL)} \\ \midrule
Cup & 0.97±0.06 (\textcolor{gggreen}{+125.6\%}) & 0.40±0.10 (\textcolor{ggdarkred}{-7.0\%}) & 0.43±0.06 & 0.30±0.20 \\ 
(w/ distractors) & 0.73±0.15 (\textcolor{gggreen}{+143.3\%}) & 0.40±0.10 (\textcolor{gggreen}{+33.0\%}) & 0.30±0.20 & 0.23±0.23 \\ \midrule
Duck & 0.63±0.21 (\textcolor{gggreen}{+57.5\%}) & 0.20±0.20 (\textcolor{ggdarkred}{-50.0\%}) & 0.40±0.17 & 0.60±0.26 \\
(w/ distractors) & 0.83±0.06 (\textcolor{gggreen}{+93.0\%}) & 0.13±0.06 (\textcolor{ggdarkred}{-69.8\%}) & 0.43±0.31 & 0.50±0.17 \\ \midrule
Pot & 0.70±0.10 (\textcolor{gggreen}{+204.3\%}) & 0.00±0.00 (\textcolor{ggdarkred}{-100.0\%}) & 0.23±0.06 & 0.53±0.11 \\ 
(w/ distractors) & 0.80±0.00 (\textcolor{gggreen}{+100.0\%}) & 0.00±0.00 (\textcolor{ggdarkred}{-100.0\%}) & 0.40±0.26 & 0.13±0.15 \\ \bottomrule
\end{tabular}
}
\label{tab:real_exp}
\end{table}

\subsection{Additional Results of Dynamics Error Evaluation}\label{app:dynamics_error}
\begin{figure}[H]
    \centering
    \includegraphics[width=\textwidth]{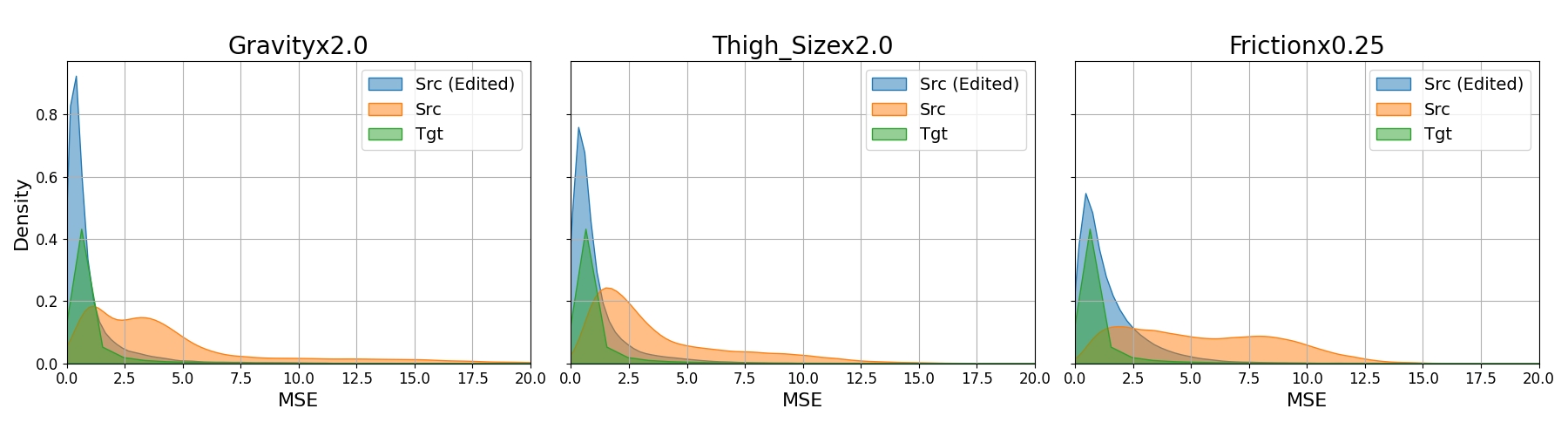}
    \caption{\small The distribution of dynamics errors (MSE) on source, edited source, and target data.}
    \label{fig:dynamics_error_mse}
\end{figure}

We have visualized the MAE distribution of dynamcis errors on original source, edited source, and target datasets. Here we report the numerical results of those dynamcis error evalutaion in Table~\ref{tab:dynamics_errors} and attach the mean square error distribution (MSE) visualized in Figure~\ref{fig:dynamics_error_mse}.

\begin{table}[H]
    \centering
    \caption{\small Dynamics errors (MSE and MAE) evaluated on original source, edited source, and target data.}
    \begin{tabular}{llcc}
        \toprule
        \textbf{Domain Gap} & \textbf{Data} & \textbf{MSE} & \textbf{MAE} \\
        \midrule
        \multirow{3}{*}{Gravity} 
            & Target           & 0.58 ± 1.75 & 0.33 ± 0.21 \\
            & Edited Source  & 1.02 ± 1.81 & 0.45 ± 0.23 \\
            & Source          & 4.62 ± 4.53 & 1.01 ± 0.44 \\
        \midrule
        \multirow{3}{*}{Thigh Size} 
            & Target           & 0.58 ± 1.75 & 0.33 ± 0.21 \\
            & Edited Source  & 1.18 ± 2.20 & 0.49 ± 0.29 \\
            & Source           & 3.88 ± 3.35 & 1.02 ± 0.46 \\
        \midrule
        \multirow{3}{*}{Friction} 
            & Target          & 0.58 ± 1.75 & 0.33 ± 0.21 \\
            & Edited Source  & 1.61 ± 2.20 & 0.61 ± 0.32 \\
            & Source           & 5.54 ± 3.36 & 1.18 ± 0.37 \\
        \bottomrule
    \end{tabular}
    \vskip -0.1in
    \label{tab:dynamics_errors}
\end{table}

\subsection{xTED with Different Choices of Editing Ratio}\label{app:t_ablation}
According to SDEdit~\citep{meng2022sdedit}, editing ratio $\kappa=0.5$ falls in the sweet spot of image editing tasks and so does it in trajectory editing tasks in general, as shown in Table~\ref{tab:t_ablation}.
$\kappa=0.0$ indicates no editing applied on the source data, whereas $\kappa=1.0$ indicates that edited source data is theoretically denoised from random noise and thus approximately equivalent to direct generation without any basis.
This reveals that $\kappa$ should not be too large or too small to effectively retain useful information from the source data while precisely eliminating domain-specific biases. 
However, in certain scenarios, $\kappa=1.0$ can successfully transform source trajectories to align with target domain properties, resulting in the best performance.
\begin{table}[H]
\centering
\caption{Ablations on editing ratio $\kappa$ with WK-MR dataset.}
\begin{tabular}[H]{clc}
\toprule
\textbf{Domain Gap} & $\kappa$ & \textbf{Target+Edited Source} \\
\midrule
\multirow{4}{*}{Gravity}   & 0.0  & 19.5±10.7  \\
                           & 0.05 & 23.1±10.4 \\
                           & 0.5  & 23.3±9.1 \\
                           & 1.0  & 27.1±10.8 \\
\midrule
\multirow{4}{*}{Friction}  & 0.0  & 17.4±6.2  \\
                           & 0.05 & 24.7±7.0  \\
                           & 0.5  & 25.9±9.1  \\
                           & 1.0  & 25.5±13.6 \\
\midrule
\multirow{4}{*}{Thigh Size} & 0.0  & 18.0±6.7  \\
                           & 0.05 & 25.7±7.9  \\
                           & 0.5  & 25.9±9.1  \\
                           & 1.0  & 24.1±9.3  \\
\bottomrule
\end{tabular}
\label{tab:t_ablation}
\end{table}

\subsection{xTED with Multiple Editing Iterations}\label{app:editing_itr}
In xTED, we add $k$-step Gaussian noise to source trajectories and denoise them in $k$-steps using a pre-trained xTED for one iteration ($e=1,\kappa=0.5$ for all our results). In Table~\ref{tab:edit_time}, we explore the impact of two editing iterations ($e=2$) and find that increasing the number of iterations can sometimes enhance xTED's performance by achieving better alignment with target domain properties. It is noteworthy that $e=1,\kappa=1.0$ and $e=2,\kappa=0.5$ involve the same total number of noising and denoising steps, with the only difference being that $e=2,\kappa=0.5$ divides the process into two iterations. 
$e=2,\kappa=0.5$ seems to perform well only if $e=1,\kappa=1.0$ also does, suggesting that some tasks with large domain gaps benefit from more editing steps. 
Interestingly, $e=2,\kappa=0.5$ always slightly outperforms $e=1,\kappa=1.0$.
This observation indicates that xTED, even with the same computational effort in noising and denoising, can achieve improved performance by dividing the editing process into finer-grained refinement stages.

\begin{table}[H]
		\centering
		\caption{xTED with different editing configurations. Averaged over 5 random seeds.}
		\adjustbox{center}{
			\begin{tabular}{clccc}
				\toprule
				Target Data & Domain Gap & $e=1,\kappa=1.0$ & $e=1,\kappa=0.5$ & $e=2,\kappa=0.5$
				\\
                \midrule

				\multirow{3}{*}{WK-MR} & Gravity & 27.1$\pm$10.8 & 23.3$\pm$9.1 & \textbf{28.3$\pm$9.2}\\
				& Friction & {25.5$\pm$13.6} & 25.9$\pm$9.1 & \textbf{26.8$\pm$8.9}  \\
                    & Thigh Size & 24.1$\pm$9.3  & \textbf{25.9$\pm$9.1} & 24.7$\pm$11.0 \\

                \bottomrule
		\end{tabular}}\label{tab:edit_time}
	\end{table}

\subsection{xTED on Larger Amount of Source Data}\label{app:data_ratio}
In this section, we present the results of leveraging a larger amount of source data (200k transitions), i.e. 10 times the target data. We tested policy performance using the TD3+BC algorithm with target data from the WK-MR and HC-MR datasets. As shown in Table~\ref{tab:ratio}, xTED demonstrates a clear advantage over the baselines and improves on the results obtained with only 20k source data. These results suggest that xTED is robust and stable in accommodating larger amounts of source data, highlighting its potential for scaling up model capacity and data usage in future developments of larger-scale cross-embodiment policy learning for robotics.

\begin{table}[H]
		\centering
            \setlength\tabcolsep{3pt}
		\caption{Average normalized scores for MuJoCo tasks on 20k target data and 200k source data.}
		\small
		\adjustbox{center}{
			\begin{tabular}{ccccccc}
				\toprule
				{Target Data} & Domain Gap & Target & Target+Source & $\Delta$ & Target+Edited Source & $\Delta$
				\\
                \midrule
				\multirow{3}{*}{\rotatebox{90}{HC-MR}}  & Gravity & 12.5$\pm$6.6 &{ {30.9$\pm$2.9} }& \textcolor{gggreen}{+17.9\%} & \textbf{33.9$\pm$2.3}& \textcolor{gggreen}{+29.4\%}\\
				& Friction & 12.5$\pm$6.6 & 26.4$\pm$3.3 & \textcolor{gggreen}{+0.8\%} & \textbf{28.0$\pm$3.2}& \textcolor{gggreen}{+6.9\%} \\
                    & Thigh Size & 12.5$\pm$6.6 &{23.7$\pm$5.4} & \textcolor{ggdarkred}{-9.5\%} & \textbf{31.0$\pm$3.7}& \textcolor{gggreen}{+18.3\%}\\
				
                    \midrule
                    \multirow{3}{*}{\rotatebox{90}{WK-MR}} & Gravity & {3.8$\pm$2.3 } & {24.5$\pm$13.6} & \textcolor{gggreen}{+47.6\%} & \textbf{{26.7$\pm$14.8}} & \textcolor{gggreen}{+60.8\%}\\
				& Friction & 3.8$\pm$2.3 & 27.9$\pm$10.8 & \textcolor{gggreen}{+68.1\%} & \textbf{{36.2$\pm$11.7}} & \textcolor{gggreen}{+118.1\%}  \\
                    & Thigh Size & 3.8$\pm$2.3 & 21.7$\pm$12.1 & \textcolor{gggreen}{+30.7\%}& \textbf{27.6$\pm$14.7} & \textcolor{gggreen}{+66.3\%}\\
                \midrule    
                & Total & 44.3 & 155.1 & \textcolor{gggreen}{+20.8\%} & \textbf{183.4} & \textcolor{gggreen}{+46.6\%}\\
    \bottomrule
		\end{tabular}}\label{tab:ratio}
	\end{table}

\subsection{Comparisons on Transformer and Temporal U-Net for Diffusion-Based Data Editing}\label{app:tranUNet}

As shown in Table~\ref{tab:tempu_ddit}, compared to Temporal U-Net commonly used in diffusion models, our architecture achieved higher overall scores and significantly reduced training time per epoch (1000 training steps).
The training time consumptions (seconds) were recorded on a server with $8\times$ A100 GPUs while each model was trained using a single GPU on the server, with no other GPUs active during the process.
The detailed architecture of Temporal U-Net can be found in~\citep{ajay2023is}.

\begin{table}[H]
\centering
\caption{Comparisons  xTED and Temporal U-Net for trajectory editing.}
\begin{tabular}{clcccc}
\toprule
\multirow{2}{*}{Target Data} & \multirow{2}{*}{Domain Gap}  & \multicolumn{2}{c}{\textbf{Temporal U-Net}} & \multicolumn{2}{c}{\textbf{xTED}} \\
\cmidrule(lr){3-4} \cmidrule(lr){5-6}
 &   & Avg. Score$\uparrow$ & Clock Time$\downarrow$ & Avg. Score$\uparrow$ & Clock Time$\downarrow$ \\
 \midrule
                     \multirow{3}{*}{HC-MR} & Gravity & 27.9$\pm$3.2 & \multirow{3}{*}{74.6s} & \textbf{31.3$\pm$2.9} & \multirow{3}{*}{\textbf{46.1s}} \\
				& Friction & 27.7$\pm$3.7 &  & \textbf{31.8$\pm$3.1} &  \\
                    & Thigh Size & 28.8$\pm$3.5 &  & \textbf{33.0$\pm$3.0} &  \\
                    \midrule
				\multirow{3}{*}{WK-MR} & Gravity & \textbf{30.9$\pm$9.6} & \multirow{3}{*}{93.2s} & 23.3$\pm$9.1 & \multirow{3}{*}{\textbf{46.5s}} \\
				& Friction & \textbf{26.3$\pm$12.6} &  & 25.9$\pm$9.1 &  \\
                    & Thigh Size & \textbf{28.2$\pm$9.3} &  & 25.9$\pm$9.1 &  \\
                    \midrule
                & Total & 169.8 & 167.8s & \textbf{171.2} & \textbf{92.6s} \\
\bottomrule
\end{tabular}
\label{tab:tempu_ddit}
\end{table}

\subsection{xTED with Additional Constraints and Conditions on Diffusion Models}\label{app:con_con}
We train xTED with constraints of fixed initial transition (Init. Const.) and no constraint of fixed last transition (Last Const.) in trajectories, like image inpainting techniques~\citep{Rombach_2022_CVPR} in previous diffusion-based image synthesis works that fix some parts of the original image during entire editing/generation process.
xTED can also accommodate conditioning strategies (Conditioned xTED).
In this section, we ablate on these additional constraints and conditions.

In terms of the implementation of return-conditioned xTED, to generate trajectories that maximize return, we condition the diffusion model on the return of trajectories, such that the training objective in Eq.~\ref{eq:denoise} should be adapted to:
\begin{equation}
    \mathcal{L}_\theta=\mathbb{E}_{k\in[1,\cdots,K],\tau_0\sim q(\tau_0),\epsilon\sim\mathcal{N}(\mathbf{0},\mathbf{I})}\|\epsilon-\epsilon_\theta(\tau_k,R(\tau),k)\|^2\label{eq:denoise_con}
\end{equation}
where the returns are normalized to ensure $R(\tau) \in [0, 1]$. 
Sampling a high-return trajectory corresponds to setting $R(\tau) \rightarrow 1$.
In the planning stage, we denoise the blurred trajectories with classifier-free guidance with perturbed noise~\citep{ajay2023is}:
\begin{equation}
    \hat{\epsilon}:=\epsilon_\theta(\tau_k,\varnothing,k)+\omega\left(\epsilon_\theta(\tau_k,R(\tau),k)-\epsilon_\theta(\tau_k,\varnothing,k)\right)\label{eq:plan_con_noise}
\end{equation}
where we assign $R(\tau)=0.9,\omega=1.0$.

As shown in Table~\ref{tab:constr_cond}, the use of initial constraint benefits the final performance, while using last constraint has the opposite effect. Notably, incorporating return guidance can further enhance the effectiveness of xTED. 

\begin{table}[H]
		\centering
		\caption{xTED with additional constraints and condition guidance. Averaged over 5 random seeds.}
		\adjustbox{center}{
			\begin{tabular}{cl|cc|cc}
				\toprule
				Target Data & Domain Gap & w/o Init. Const. & w/ Last Const. & xTED & Conditioned xTED
				\\
                \midrule
                    \multirow{3}{*}{HC-MR} & Gravity & 28.6$\pm$4.3 & 21.8$\pm$3.3 & \textbf{31.3$\pm$2.9} & 30.5$\pm$3.1 \\
				& Friction & 26.3$\pm$4.1 & 22.0$\pm$3.7 & 31.8$\pm$3.1 & \textbf{35.1$\pm$1.8} \\
                    & Thigh Size & 29.6$\pm$3.4 & 21.3$\pm$3.9 & 33.0$\pm$3.0 & \textbf{33.3$\pm$3.0} \\
                    \midrule
				\multirow{3}{*}{WK-MR} & Gravity &\textbf{ 24.5$\pm$7.7 }& 18.7$\pm$6.9 & 23.3$\pm$9.1 & \textbf{24.5$\pm$9.0}\\
				& Friction &{ 20.6$\pm$7.5 }& 15.2$\pm$6.7 & \textbf{25.9$\pm$9.1} & 22.3$\pm$8.7  \\
                    & Thigh Size & 18.2$\pm$7.3 & 14.5$\pm$7.0 & 25.9$\pm$9.1 & \textbf{31.1$\pm$11.5} \\
                    \midrule
                & Total & 147.8 & 113.5 & 171.2 & \textbf{176.8}\\
                \bottomrule
		\end{tabular}}\label{tab:constr_cond}
	\end{table}

\subsection{xTED with Different Choices of Trajectory Horizons for Diffusion Modeling}\label{app:horizon}
In sequential modeling, the sequence horizon is a critical parameter. Modeling sequences that are too long can significantly challenge the model capacity, while sequences that are too short often result in reduced long-horizon consistency. In this section, we compare a shorter horizon of 10 with the horizon of 20 used in our main simulation experiments. As shown in Table~\ref{tab:horizon}, selecting a horizon of 20 improves the model performance in the HalfCheetah and Walker2d locomotion tasks.


\begin{table}[H]
		\centering
            \setlength\tabcolsep{3pt}
		\caption{Average normalized scores for MuJoCo tasks on horizon choices for trajectory modeling.}
		\small
		\adjustbox{center}{
			\begin{tabular}{ccccc}
				\toprule
				\multicolumn{2}{c}{Target Data} & Domain Gap & Target+Edited Source (\textbf{horizon=20})& Target+Edited Source (\textbf{horizon=10})
				\\
                \midrule
				\multirow{9}{*}{\rotatebox{90}{Halfcheetah}} &\multirow{3}{*}{\rotatebox{90}{Med}} & Gravity & \textbf{40.6$\pm$2.0}& 40.1$\pm$2.4\\
				&& Friction & \textbf{41.2$\pm$1.6}& 40.6$\pm$1.6 \\
                    && Thigh Size & \textbf{40.7$\pm$2.4}& 40.7$\pm$2.2\\
                    \cmidrule(lr){2-5}
				&\multirow{3}{*}{\rotatebox{90}{Med-R}} & Gravity & \textbf{31.3$\pm$2.9} & 28.8$\pm$3.3 \\
				&& Friction & \textbf{31.8$\pm$3.1} & 29.8$\pm$2.8 \\
                    && Thigh Size & \textbf{33.0$\pm$3.0} & 30.2$\pm$3.2\\
                    \cmidrule(lr){2-5}
				&\multirow{3}{*}{\rotatebox{90}{Med-E}} & Gravity & \textbf{43.8$\pm$3.6} & 43.6$\pm$3.6 \\
				&& Friction & \textbf{43.2$\pm$3.0} & 37.6$\pm$3.8 \\
                    && Thigh Size &\textbf{43.0$\pm$3.0} & 40.6$\pm$3.5\\
                    \midrule
                    \midrule
                    \multirow{9}{*}{\rotatebox{90}{Walker2d}}&\multirow{3}{*}{\rotatebox{90}{Med}} & Gravity & \textbf{{58.2$\pm$11.7}} & 53.1$\pm$12.6\\
				&& Friction & 54.5$\pm$13.7 & \textbf{58.9$\pm$12.5}  \\
                    && Thigh Size & \textbf{58.9$\pm$11.7} & 57.8$\pm$12.0\\
                   \cmidrule(lr){2-5}
				&\multirow{3}{*}{\rotatebox{90}{Med-R}} & Gravity & \textbf{23.3$\pm$9.1} & 21.8$\pm$7.1\\
				&& Friction & \textbf{25.9$\pm$9.1} & 25.0$\pm$9.1 \\
                    && Thigh Size & \textbf{25.9$\pm$9.1} & 18.9$\pm$7.5 \\
                    \cmidrule(lr){2-5}
				&\multirow{3}{*}{\rotatebox{90}{Med-E}} & Gravity & \textbf{82.9$\pm$18.1} & 76.0$\pm$20.0 \\
				&& Friction &\textbf{74.0$\pm$24.5} & 73.2$\pm$16.2 \\
                    && Thigh Size & \textbf{{81.0$\pm$21.4}} & 75.1$\pm$22.9\\
                    \midrule
                && Total & \textbf{833.2} & 791.8\\
    \bottomrule
		\end{tabular}}\label{tab:horizon}
	\end{table}
 
\subsection{xTED on Different Degrees of Dynamic Gaps}\label{app:degree_gap}

\begin{figure}[H]
    \centering
    \includegraphics[width=0.5\textwidth]{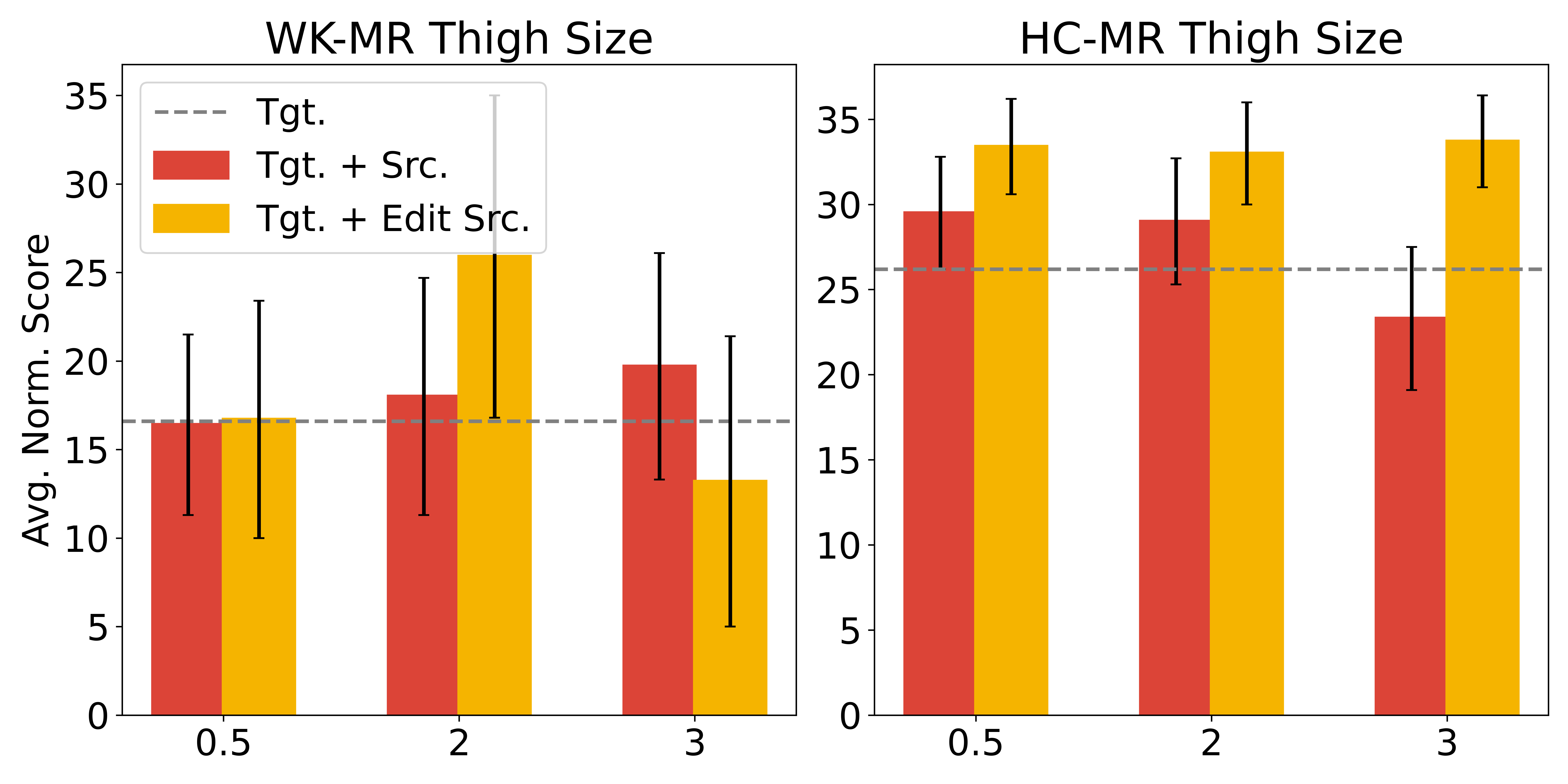}
    \caption{\small Average normalized returns for different degrees of dynamic gaps (different thigh sizes) on 20k transitions from WK-MR and HC-MR.}
    \label{fig:data_gap}
\end{figure}

We explore the impact of different degrees of dynamics gap on policy performance aided with xTED. Specifically, we modify the thighs of HalfCheetah and Walker2d to $0.5\times$, $2\times$, and $3\times$ original sizes respectively. 
The results in Fig.~\ref{fig:data_gap} indicate that under various degrees of dynamics gaps, 
edited source data can augment the original dataset effectively in most of situations, while directly augmenting with unedited source data is sometimes more sensitive to extreme domain gaps in HalfCheetah environments.

\subsection{xTED for Multiple Source Domains with Different Gaps}\label{app:multi-source}
xTED demonstrates the ability to handle diverse source domains with varying domain gaps using a single pre-trained diffusion model over the target domain, without the need for re-training or fine-tuning. 
We integrate source datasets exhibiting three distinct domain gaps (20k data with thigh size, gravity, and friction gaps, respectively) into the policy learning process. As shown in Table~\ref{tab:my_label}, while unedited source data with intricate domain gaps generally degrade overall policy performance, the edited source data, with narrower gaps, consistently improves policy learning outcomes.

\begin{table}[H]
    \centering
    \setlength\tabcolsep{3pt}
    \caption{Average normalized scores for multi-source MuJoCo experiments.}
    \small
    \adjustbox{center}{
    \begin{tabular}{cc|cc|cc}
    \toprule
         Target Data  & Tgt & Tgt+Src & $\Delta$ & Tgt+Src(Edited) & $\Delta$\\
         \midrule
         HC-M  & 39.5$\pm$2.4 & 37.6$\pm$3.1 & \textcolor{ggdarkred}{-4.8\%} & \textbf{39.7$\pm$1.9} & \textcolor{gggreen}{+0.5\%} \\
         HC-MR  & 26.2$\pm$3.5 & 26.6$\pm$3.5 & \textcolor{gggreen}{+1.5\%} & \textbf{28.2$\pm$3.2} & \textcolor{gggreen}{+7.6\%} \\
         HC-ME  & 40.1$\pm$2.9 & 35.3$\pm$5.0 & \textcolor{ggdarkred}{-12.0\%} & \textbf{41.3$\pm$1.8} & \textcolor{gggreen}{+3.0\%} \\
         \midrule
         WK-M  & 45.3$\pm$15.9 & 41.8$\pm$19.1 & \textcolor{ggdarkred}{-7.7\%} & \textbf{58.5$\pm$9.3} & \textcolor{gggreen}{+29.1\%} \\
         WK-MR  & 16.6$\pm$5.9 & 19.8$\pm$6.2 & \textcolor{gggreen}{+19.3\%} & \textbf{22.7$\pm$7.6} & \textcolor{gggreen}{+36.7\%} \\
         WK-ME  & 71.0$\pm$21.0 & 74.2$\pm$15.2 & \textcolor{gggreen}{+4.5\%} & \textbf{81.3$\pm$18.0} & \textcolor{gggreen}{+14.5\%} \\
         \midrule
         Total  & 238.7 & 235.3 & \textcolor{ggdarkred}{-1.4\%} & \textbf{271.7} & \textcolor{gggreen}{+13.8\%} \\
    \bottomrule
    \end{tabular}}
    \label{tab:my_label}
\end{table}

\subsection{Frequency-Domain Analysis of Trajectory Editing}\label{app:freq}

To elucidate the mechanism by which xTED corrects domain gaps while preserving task-relevant semantics, we conduct a frequency-domain analysis of trajectory differences, drawing an analogy to image editing techniques like SDEdit~\citep{meng2022sdedit} which preserve core structures via partial noising. In our Mujoco locomotion experiments, we randomly sample 100 transitions from source datasets (with gaps in gravity, friction, and thigh size) and target datasets (HCMR) to compute the Fast Fourier Transform (FFT). By comparing magnitude differences across normalized frequency components, as shown in Table~\ref{tab:frequency_analysis}, we observe that domain gaps are primarily concentrated in the medium-to-high frequency range ($0.2$--$0.5$), whereas low-frequency components ($\leq 0.2$)—representing structural task semantics like forward movement or jumping—remain remarkably consistent across domains. Similar findings are also revealed in previous works on diffusion models~\citep{DDPM, huang2025fresca}. Leveraging the property of diffusion models where the forward process masks high-frequency features first and the denoising process reconstructs them according to the training data distribution, xTED effectively erases domain-specific discrepancies in high-frequency bands. During denoising, the target-trained model reconstructs these bands using target-specific physical patterns while scaffolding on the preserved low-frequency task structures, thereby facilitating effective policy learning in the target domain.

\begin{table}[htbp]
  \centering
  \caption{Magnitude Differences of Trajectories in the Frequency Domain.}
  \label{tab:frequency_analysis}
  \small
  \begin{tabular}{ccccc}
    \toprule
    \textbf{Normalized} & \textbf{Mag. Diff.} & \textbf{Mag. Diff.} & \textbf{Mag. Diff.} & \textbf{Mag. Diff.} \\
    \textbf{Frequency} & \textbf{[HCM-HCMR(Baseline)]} & \textbf{[Src.(Gravity)-Tgt.(HCMR)]} & \textbf{[Src.(Friction)-Tgt.(HCMR)]} & \textbf{[Src.(Thigh Size)-Tgt.(HCMR)]} \\
    \midrule
    0.05 & 5.57
    & 7.30   & 6.24  & 4.18  \\
    0.10 & 6.82  & 6.12   & 7.62  & 7.55  \\
    0.15 & 20.00 & 14.63  & 16.61 & 28.87 \\
    0.20 & 16.96 & \textbf{172.65} & 16.19 & \textbf{88.68} \\
    0.25 & 11.54 & \textbf{20.15}  & \textbf{33.17} & \textbf{24.38 }\\
    0.30 & 12.64 & 11.91 & \textbf{25.35} & \textbf{15.28} \\
    0.35 & 11.31 & 11.62  & \textbf{15.77} & \textbf{16.38} \\
    0.40 & 5.35  & \textbf{21.31}  & 8.59  & 6.46  \\
    0.45 & 8.93  & 10.77  & 9.78  & 6.50  \\
    \bottomrule
  \end{tabular}
\end{table}

\end{document}